%% file: ARXIV_nv_report_eola.tex
\pdfoutput=1
\documentclass[10pt, logo, onecolumn, copyright]{nvidiatechreport}

\usepackage{mdframed}
\usepackage[utf8]{inputenc} 
\usepackage[T1]{fontenc}    

\usepackage{hyperref}
\usepackage{url}
\usepackage{tabularx}
\usepackage{booktabs}
\usepackage{graphicx}
\usepackage{multirow}
\usepackage{array}
\usepackage{algorithm}
\usepackage{algpseudocode}
\usepackage{wrapfig,lipsum}
\usepackage{graphicx}
\usepackage{subfig}
\usepackage[dvipsnames]{xcolor}
\usepackage{amsmath, amsthm, amssymb}
\usepackage{enumitem}
\usepackage{natbib}
\usepackage{float}

\newcommand{\newproblem}{customized compensation}
\newcommand{\methodabbr}{EoRA}

\newcommand{\methodbf}{fine-tuning-free \textbf{E}igenspace L\textbf{o}w-\textbf{R}ank \textbf{A}pproximation}
\newcommand{\losscompensation}{error approximation loss}

\DeclareMathOperator*{\argmin}{arg\,min}

\usepackage{fp} 

\usepackage{xcolor}

\title{EoRA: Fine-tuning-free Compensation for Compressed LLM with Eigenspace Low-Rank Approximation}

\author{Shih-Yang Liu\textonesuperior, Maksim Khadkevich, Nai Chit Fung\textonesuperior, Charbel Sakr, 
Chao-Han Huck Yang, Chien-Yi Wang, Saurav Muralidharan, Hongxu Yin, 
Kwang-Ting Cheng\textonesuperior, Jan Kautz, \quad\quad\quad\quad\quad\quad\quad\quad\quad\quad
Yu-Chiang Frank Wang, Pavlo Molchanov, Min-Hung Chen}

\correspondingauthor{X}

\begin{abstract}

\textbf{Abstract:} While post-training compression techniques effectively reduce the memory footprint, latency, and power consumption of Large Language Models (LLMs), they often result in noticeable accuracy degradation and remain limited by hardware and kernel constraints that restrict supported compression formats—ultimately reducing flexibility across a wide range of deployment scenarios. In this work, we propose EoRA—a novel, \textbf{fine-tuning-free} method that augments compressed LLMs with low-rank matrices, allowing users to rapidly enhance task-specific performance and freely balance the trade-off between accuracy and computational overhead beyond the constraints of compression formats. \methodabbr~consistently outperforms prior fine-tuning-free low-rank methods in recovering the accuracy of compressed LLMs, achieving notable accuracy improvements
(e.g., $\mathbf{10.84\%}$ on ARC-Challenge, $\mathbf{6.74\%}$ on MathQA, and $\mathbf{11.45\%}$ on GSM8K for LLaMA3-8B compressed to 3-bit). We also introduce an optimized CUDA kernel, accelerating inference by up to 1.4× and reducing memory overhead through quantizing EoRA. Overall, \methodabbr~offers a prompt solution for improving the accuracy of compressed models under varying user requirements, enabling more efficient and flexible deployment of LLMs. Code is available at \url{https://github.com/NVlabs/EoRA}.
\end{abstract}

\begin{document}
\maketitle

\noindent\textbf{Links:
\href{https://github.com/NVlabs/EoRA}{NVLabs Code},
\href{https://github.com/ModelCloud/GPTQModel/tree/main/examples/eora}{GPTQModel Support} | 
\href{https://developer.nvidia.com/blog/a-fine-tuning-free-approach-for-rapidly-recovering-llm-compression-errors-with-eora/}{NV Tech Blog}
}

\input{EDIT_ME_eola_main}

\end{document}

%% file: EDIT_ME_eola_main.tex
\begin{figure}[h]
\begin{center}
\includegraphics[width=0.85\columnwidth]{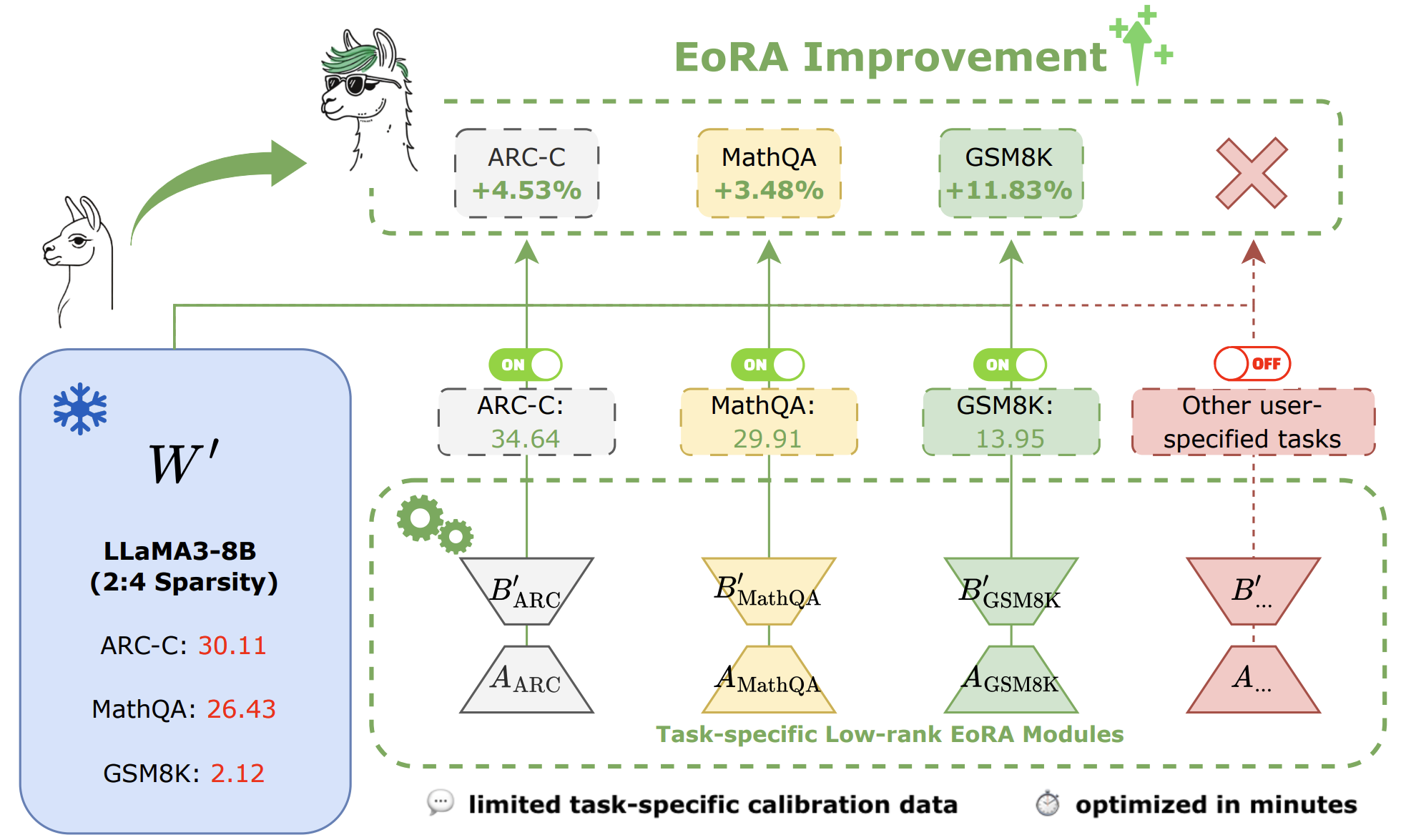}
\end{center}
\caption{
An overview of our proposed \methodabbr, which enables swift task-specific accuracy enhancement for compressed LLMs without \textbf{fine-tuning}, using only a small amount of downstream calibration data. At inference time, a single compressed backbone is loaded, while lightweight, task-specific low-rank modules can be dynamically toggled on and off on demand, enabling efficient and flexible deployment. \methodabbr~with rank 128 boosts the accuracy of the LLaMA3-8B model pruned to $2{:}4$ structured sparsity by $4.53\%$, $3.48\%$, and $11.83\%$ on ARC-C, MathQA, and GSM8K, respectively—all achieved within minutes using just 64 calibration samples per task.}
\vskip -0.2in
\label{teaser-method}
\end{figure}

\section{Introduction}

Although Large Language Models (LLMs) excel in various tasks, their deployment remains challenging due to high inference costs. Post-training compression methods, like quantization~\citep{frantar2022gptq, lin2024awq, liu2024spinquant, tseng2024quip} and pruning~\citep{ma2023llm, frantar2023sparsegpt, sun2023simple}, aim to reduce computational demands but typically cause accuracy loss or face hardware/kernel constraints,  limiting deployment flexibility. For instance, strict hardware-supported formats, such as 2:4 sparsity on NVIDIA GPUs or integer-only quantization kernels, prevent intermediate approaches (e.g., 2.X:4 sparsity or arbitrary-bit quantization) that could offer a more adaptable trade-off between accuracy and latency based on user needs.

To relax these format constraints and improve the accuracy of the compressed models on specified tasks, we formulate a new problem, termed \textit{\newproblem}: Given a compressed LLM, we attach residual low-rank paths to it to \textit{compensate} for compression errors and enhance task-specific accuracy, enabling more flexible control over the trade-off between accuracy and compression ratio to accommodate varying user requirements. For example, a user may wish to boost the accuracy of a 2:4 sparsity-pruned model on math reasoning tasks, accepting a modest increase in memory usage and inference latency in return. Importantly, in our problem setting, the weights of the compressed model are not modified during compensation. This enables deployment of a single, general compressed backbone alongside lightweight, task-specific low-rank modules that can be dynamically loaded as needed—allowing for efficient integration with existing multi-adapter inference frameworks (e.g., vLLM~\citep{multilora}) as illustrated in Figure~\ref{teaser-method}. A naive solution is to apply SVD~\citep{liloftq,yao2024zeroquantv2} for compensation; however, this neglects calibration data and thus fails to enhance task-specific performance. Alternatively, LoRA-based methods, such as \cite{liloftq,dettmers2024qlora}, require fine-tuning, limiting their applicability for rapid task adaptation. These limitations prompt an important question: “\textit{How can we swiftly improve the task-specific accuracy for compressed LLMs without fine-tuning?}”

To tackle this research challenge, we introduce \textit{\methodbf~(\textbf{\methodabbr})}, a method designed to efficiently enhance the task-specific accuracy of compressed LLMs while offering users greater flexibility in managing the trade-off between accuracy and computational overhead. \methodabbr~operates by projecting the compression error into the task-specific eigenspace of each layer’s input activations, followed by applying SVD to approximate the projected error. This approach ensures that the SVD approximation error directly aligns with the task-specific compression loss. As a \textbf{fine-tuning-free} method, \methodabbr~avoids backpropagation and completes in just a few minutes using minimal calibration data. 

We validate the effectiveness of \methodabbr~in boosting the accuracy of compressed LLMs (LLaMA2-7B/13B and LLaMA3-8B) on language generation, commonsense reasoning, and math tasks. Our method consistently outperforms other fine-tuning-free baselines, especially for aggressively compressed (including \textit{pruned, quantized, and both}) models (e.g., $\mathbf{2.65\%}$, $\mathbf{3.42\%}$, and $\mathbf{10.99\%}$ improvement over ZeroQuant-V2 on ARC-Challenge, MathQA, and GSM8K when compensating 2:4 pruned LLaMA3-8B). To reduce redundant memory transfer overhead from running low-rank compensation, we design a fused kernel that integrates low-rank and quantization operations, achieving up to 1.4× speedup.

The summary of our contributions is as follows:
\begin{itemize}[leftmargin=*]
\item \textbf{Flexible and Task-specific Model Compensation}: We propose, \textit{\methodbf~(\textbf{\methodabbr})}, a \textcolor{Green}{\emph{fine-tuning-free}} approach that improves the task-specific accuracy of compressed LLMs \textcolor{Green}{\emph{in minutes}} using \textcolor{Green}{\emph{minimal calibration data}}, while supporting more flexible compression ratios unconstrained by hardware or kernel-imposed format limitations.

\item \textbf{Eigenspace Projection}: \methodabbr~leverages calibration data to project the compression error into the task-specific eigenspace and utilizes the corresponding eigenvalues as importance indicators, effectively aligning the approximation error with task-specific compression loss.

\item \textbf{Efficient Inference}: We develop a custom kernel that fuses part of the low-rank matrix multiplication with a quantization kernel, accelerating \methodabbr~inference by up to 1.4x. \methodabbr~is also robust to quantization, further minimizing the size-overhead from low-rank compensation matrices.
\end{itemize}

\newpage
\section{Preliminaries: Post-training Compression}
\label{sec:preliminaries}

\textbf{Post-training compression} aims to compress a well-trained model by a targeted compression ratio, utilizing only a limited set of calibration data. The compression process is often framed as a layer-wise optimization problem, aiming to minimize the layer-wise output difference between the original weight $W_l \in \mathbb{R}^{d \times k}$ and the compressed weight $\hat{W}_l \in \mathbb{R}^{d \times k}$ for each layer $l$.  
Then the \textit{layer-wise model compression loss} can be formed as: 
{\small
\begin{equation}
    \argmin_{\hat{W}_l} ||W_{l}X_{l} - \hat{W}_lX_{l}||_F 
\label{eq:layer_objective}
\end{equation}}
where $X_{l} \in \mathbb{R}^{k \times n}$ is the input activation of layer $l$ and $F$ denotes the Frobenius error between the layer-wise output. Once the compression is complete, the $W_l$ for each layer will be substituted with $\hat{W}_l$, resulting in a smaller model size, faster inference, or both. However, their flexibility is often limited by a discrete set of compression formats (e.g., 2:4 sparsity, 3/4-bit quantization), making it challenging to meet the diverse accuracy/overhead requirements of different users.

To bypass the limitations of fixed compression formats and enhance the accuracy of compressed models on user-specified tasks, we introduce a new problem, termed \textit{\newproblem}: Given an already compressed model, the objective is to add residual low-rank paths that \textit{compensate} for compression errors and enhance task-specific accuracy according to user-defined accuracy/overhead requirements. Crucially, the compressed model’s weights remain unchanged during compensation, enabling the deployment of a single, general compressed backbone with lightweight, task-specific low-rank modules that can be dynamically loaded as needed, facilitating efficient integration with existing inference frameworks, as illustrated in Figure~\ref{teaser-method}.

A simple approach to obtain low-rank residual paths that compensate for compression errors is to directly apply Singular Value Decomposition (SVD)~\citep{liloftq,yao2024zeroquantv2,li2024svdqunat}. More specifically, this method relies on a closed-form solution by using SVD to approximate the compression error $\Delta W_l=W_l-\hat{W}_l$ as $\Delta W_l \approx U_l \Sigma_l V_l^T $, where $\Sigma_l \in \mathbb{R}^{r \times r}$ is a diagonal matrix containing the top-$r$ largest singular value sorted in descending order, and $U_l \in \mathbb{R}^{d \times r}$, $ V_l \in \mathbb{R}^{k \times r}$ are orthonormal matrices, with each column representing the singular vectors corresponding to the singular values in $\Sigma_l$. The product of $ U_l $ and $ \Sigma_l$ can then be treated as $B_l = U_l\Sigma_l $ with $V_l^T$ being treated as $A_l$. Overall, the \textit{\losscompensation} can be formulated as:
{\small
\begin{equation}
    \argmin_{B_l,A_l} ||\Delta W_l - B_{l}A_{l}||_F 
\label{eq:weight_objective}
\end{equation}}
and SVD is applied on $\Delta W_l$ to minimize the above equation. 
However, naively applying SVD to optimize \losscompensation~(Eq.\ref{eq:weight_objective}) does not ensure minimization of the layer-wise compression loss (Eq.\ref{eq:layer_objective}) and ignores calibration data, making it ineffective for task-specific accuracy recovery. While LoRA-based methods~\citep{liloftq,dettmers2024qlora} address this issue, they require fine-tuning and are less suitable for rapid adaptation. This raises a key question: “\textit{How can we swiftly improve the task-specific accuracy for compressed LLMs without fine-tuning?}”. For simplicity, we omit the subscript $l$, which corresponds to layer $l$ in the following sections.

\section{Method: \methodabbr}
\label{sec:method}

To tackle the challenge of improving task-specific accuracy of compressed LLMs without fine-tuning, we introduce \textit{\methodbf~(\textbf{\methodabbr})}—a method that preserves the efficiency of existing fine-tuning-free solutions while substantially improving their \textit{effectiveness} in task-specific accuracy recovery.

First, we propose projecting the compression error into the eigenspace~\citep{stewart2001matrix} of the corresponding layer's input activations, ensuring a direct alignment between the \losscompensation~(Eq.~\ref{eq:weight_objective}) and the overall layer-wise model compression loss (Eq.~\ref{eq:layer_objective}). Inspired by the classical Principal Component Analysis (PCA) algorithm, we leverage the eigenvalues of each activation channel as importance scores to indicate the importance of each column after the eigenprojection. This allows us to allocate more low-rank representation capacity to approximate the more critical error elements.
Following PCA, we perform the eigendecomposition on $\Tilde{X}\Tilde{X}^T$ where $\Tilde{X} \in \mathbb{R}^{k \times n}$ is the average of the input activations over the task-specific calibration set. The eigendecomposition $\Tilde{X}\Tilde{X}^T = Q \Lambda Q^{T}$ is then used to derive the eigenspace projection matrix $Q \in \mathbb{R}^{k \times k}$, whose columns are the eigenvectors, and $\Lambda \in \mathbb{R}^{k \times k}$, which is a diagonal matrix with each diagonal element being the corresponding eigenvalues of the eigenvectors in $Q$.
We then propose to project the compression error $\Delta W$ into the eigenspace with the projection matrix $Q' = Q \sqrt{\Lambda}$ to obtain the projected error $\Delta W'\in \mathbb{R}^{d \times k} = \Delta W Q'$. The proposed new \losscompensation, \textit{\methodabbr~loss}, can be formulated as:
{\small
\begin{equation}
   \argmin_{B',A'} ||\Delta W' - B'A'||_F 
\label{eq:overall_eq}
\end{equation}}
where SVD is applied to approximate $\Delta W'$ as $\text{SVD}(\Delta W') \approx U'\Sigma'V'^T$, and $\Sigma' \in \mathbb{R}^{r \times r}$ contains the top-$r$ singular values. $U' \in \mathbb{R}^{d \times r}$ and $V' \in \mathbb{R}^{k \times r}$ are orthonormal matrices with columns representing the corresponding singular vectors. Then the low-rank matrices $B'$ and $A'$ are then assigned as $B' = U'  \Sigma'$ and $A' = V'^T$. This loss function ensures that error columns associated with larger eigenvalues are approximated more accurately than those with smaller eigenvalues. We then multiply the low-rank approximation in the eigenspace $\Delta W'$ with $ Q'^{-1}=\sqrt{\Lambda}^{-1}Q^T$ to project back to the original space, obtaining the final task-specific compression error approximation as $\Delta W = \Delta W'Q'^{-1} \approx B'A'Q'^{-1}$. $Q'$ is invertible because $Q'^{-1} = \sqrt{\Lambda}^{-1} Q^{T}$, and $Q'Q'^{-1} = Q \sqrt{\Lambda}  \sqrt{\Lambda}^{-1} Q^{T}$. Here, the middle term $\sqrt{\Lambda}  \sqrt{\Lambda}^{-1}$ simplifies to the identity matrix, and since $Q$ is an orthogonal matrix, $QQ^T$ also yields the identity matrix.
The product of $A'$ and $Q'^{-1}$ can be consolidated into a single matrix with the same dimensions as the original $A'$, ensuring no additional inference latency as $A = A'Q'^{-1}$. Then, the forward pass of one linear layer of the compressed model compensated with \methodabbr~for the input activation $X$ can be formulated as: 
{\small
\begin{equation}
\hat{W}X + B'AX
\end{equation}}

EoRA compensation is applied to each compressed linear layer, and the overall \textbf{fine-tuning-free} optimization of Eq.~\ref{eq:overall_eq} across all linear layers can be completed in just a few minutes, enabling users to rapidly enhance the accuracy of compressed LLMs on their chosen downstream tasks using only a small amount of task-specific calibration data—without any need for backpropagation. \methodabbr~can also provide better initialization for further LoRA fine-tuning, offering users the option to further improve accuracy if additional computational resources are available. Moreover, the low-rank matrices of \methodabbr~are robust to quantization, which can further reduce the additional memory/inference cost. Please refer to Sec.~\ref{model-size} for more details.

\begin{algorithm}
\footnotesize
\caption{Eigenspace
low-rank approximation (\methodabbr)}\label{alg:trex}
\begin{algorithmic}
\State \textbf{Input:} $\Tilde{X}$: Average of the input activations of the current layer over the calibration set, $W$: Full-precision Weight, $\hat{W}$: Compressed Weight, $r$: Compensation rank 
\State \textbf{Output:} $B',A$: Two low-rank matrices for compensation.
\State 1. $\Delta W = W - \hat{W}$
\State 2. Run Eigendecompostion on $\Tilde{X}\Tilde{X}^T = Q \Lambda Q^{T}$
\State 3. Reformulate $Q \Lambda Q^{T} =  (Q \sqrt{\Lambda})(\sqrt{\Lambda}Q^T) = Q'Q'^T $
\State 4. Project the compression error to eigenspace $\Delta W'= \Delta W Q'$
\State 5. Run $r$-rank SVD approximation on $\Delta W'$, $ B'A' = U' \Sigma' V' = \text{SVD}(\Delta W')$
\State 6. Project the approximation back to the original space $A = A'Q'^{-1}$
\State 7. The final forward pass of current layer becomes $\hat{W}X + B'AX$
\end{algorithmic}
\end{algorithm}

\paragraph{Mapping \methodabbr~loss~(Eq.~\ref{eq:overall_eq}) to task-specific compression loss~(Eq.~\ref{eq:layer_objective}):}
When Eq.~\ref{eq:layer_objective} is conditioned on different task-specific calibration data, it also implies the compressed model’s accuracy on each corresponding task. Therefore, the objective of task-specific low-rank compensation is to approximate $\Delta W$ that minimizes Eq.~\ref{eq:layer_objective}, using input activations $X$ derived from the calibration data of different tasks. To achieve this, we reformulate the compression objective for each layer as:
{\small
\begin{equation}
    \argmin_{B,A} || WX - (\hat{W}+BA)X||_F = \argmin_{B,A} ||\Delta WX - BAX||_F
\label{eq:layer_objective_rewrite}
\end{equation}}
Since the Frobenius norm of a matrix is equal to the square root of its Gram matrix \citep{sun1991proof, wang2024svdllm}, the minimization problem can be rewritten as: 
{\small
\begin{flalign}
\begin{aligned}
&\argmin_{B,A} ||\Delta WX - BAX||_F =\argmin_{B,A} [ \text{trace}((\Delta W-BA)XX^T(\Delta W-BA)^T) ]^{\frac{1}{2}}
\label{eq:layer_objective_rewrite_2}
\end{aligned}
\end{flalign}}

Directly applying SVD on $\Delta W$ initially does not guarantee the minimization of the above equation Eq.~\ref{eq:layer_objective_rewrite_2}. To address this issue, \methodabbr~projects $\Delta W$ into the eigenspace before performing SVD. In the following, we demonstrate that minimizing Eq.~\ref{eq:overall_eq} with SVD is the same as minimizing Eq.~\ref{eq:layer_objective_rewrite_2}.

\textbf{Theorem 1.} \textit{For an activation matrix $X$, whose matrix product $XX^T$ has an eigendecomposition given by $XX^T = Q\Lambda Q^T$. By projecting the compression error $\Delta W$ into the eigenspace with $Q\sqrt{\Lambda}$ as $\Delta W' = \Delta W Q\sqrt{\Lambda}$, minimizing Eq.~\ref{eq:overall_eq} via SVD becomes equivalent to minimizing Eq.~\ref{eq:layer_objective_rewrite_2}.}

\textit{Proof.} First, note that $XX^T = Q\Lambda Q^T$, and by substituting this into Eq.~\ref{eq:layer_objective_rewrite_2}, we get 
{\small
\begin{flalign}
\begin{aligned}
&[ \text{trace}((\Delta W-BA)Q\Lambda Q^T(\Delta W-BA)^T) ]^{\frac{1}{2}} \\
& = [ \text{trace}((\Delta WQ-BAQ)\Lambda (\Delta WQ-BAQ)^T) ]^{\frac{1}{2}}
\label{eq:layer_objective_rewrite_3}
\end{aligned}
\end{flalign}}
Since $\Lambda = \sqrt{\Lambda}\sqrt{\Lambda}$ and $\sqrt{\Lambda} = \sqrt{\Lambda}^{T}$, the above Eq.~\ref{eq:layer_objective_rewrite_3} can further be rewritten as:
{\small
\begin{flalign}
\begin{aligned}
\relax [ \text{trace}((\Delta WQ\sqrt{\Lambda}-BAQ\sqrt{\Lambda})(\Delta WQ\sqrt{\Lambda}-BAQ\sqrt{\Lambda})^T) ]^{\frac{1}{2}}
\label{eq:layer_objective_rewrite_4}
\end{aligned}
\end{flalign}}
Let $Q' = Q\sqrt{\Lambda}$, then Eq.~\ref{eq:layer_objective_rewrite_4} becomes:
{\small
\begin{flalign}
\begin{aligned}
&[ \text{trace}((\Delta W Q'-BA Q')(\Delta W Q'-BAQ')^T) ]^{\frac{1}{2}} \\
& = [ \text{trace}((\Delta W' - BA Q')(\Delta W' - BAQ')^T) ]^{\frac{1}{2}} \\
& = ||\Delta W' - BA Q'||_F
\label{eq:layer_objective_rewrite_5}
\end{aligned}
\end{flalign}}where the square root of the Gram matrix can be transformed back to the corresponding Frobenius norm according to \citep{sun1991proof}. By setting $BAQ' = B'A'$, $||\Delta W' - BA Q'||_F$ becomes $||\Delta W' - B'A'||_F$. By the Eckart–Young theorem \citep{Eckart_Young_1936}, the minimization of this Frobenius norm is achieved by running SVD on $\Delta W'$, therefore, we prove that minimizing $||\Delta W' - B'A'||_F$ via SVD is equivalent to minimizing Eq.~\ref{eq:layer_objective_rewrite_2}, where low-rank approximation of $\Delta W'$ is $ \text{SVD}(\Delta W') = B'A'$. Note that the above minimization is constrained to the rank of $A'$ and $B'$.

\section{Experiments}
\subsection{Experiments Details}
\label{sec:experiments_details}
We implement \methodabbr~in PyTorch~\citep{paszke2017automatic}, utilizing the Hugging Face Transformers and Datasets framework~\citep{wolf2019huggingface}. All experiments are conducted on a single NVIDIA H100 GPU. We primarily focus on evaluating \methodabbr~for compensating LLaMA2-7B/13B and LLaMA3-8B models, compressed using SparseGPT~\citep{frantar2023sparsegpt}, a widely adopted pruning method, and GPTQ~\citep{frantar2022gptq} for quantization. Channel-wise asymmetric quantization is applied across all experiments, and we follow the settings from \citep{huang2024good} to construct the calibration dataset for both SparseGPT and GPTQ. 

We compare \methodabbr~with \textbf{ZeroQuant-V2}~\citep{yao2024zeroquantv2} which proposes using simple SVD for optimizing Eq.~\ref{eq:weight_objective}. Although Activation-aware Singular Value Decomposition (ASVD)~\citep{yuan2023asvd} is designed to replace the entire model with its low-rank decomposition rather than approximating the compression errors, its strategy of incorporating activation distribution variance can also be adapted for error compensation using low-rank matrices. Specifically, we scale the compression error $\Delta W$ using a diagonal scaling matrix $S$, where each diagonal entry $S_{ii}$ is computed based on the average absolute value of the activations $\tilde{X}$ in the $i$-th channel as $S_{ii} = \left(\frac{1}{n}\sum_{j=1}^{n} |\tilde{X}_{ij}|\right)^{\frac{1}{2}}$. Here, $n$ denotes the number of activation entries in the $i$-th channel. We then apply SVD to the scaled error $\Delta W^{"} = \Delta W S$ to obtain its low-rank approximation. Since $S$ is invertible, we can project the approximation back to the original space as $\Delta W = \Delta W^{"} S^{-1} \approx B^{"}A^{"}S^{-1}$—same as how EoRA project its compensation back to the original space. We refer to this method as \textbf{Act-S} in the remainder of this paper. We also compare EoRA with a training-based method, \textbf{ApiQ}~\citep{liao2024apiq}, which optimizes low-rank matrices ($A$ and $B$) using gradient-based training to minimize Eq.~\ref{eq:layer_objective_rewrite_2}. In our comparison, we limit the evaluation to the layer-wise variant of ApiQ, as other variants require substantially more memory or training time. These more resource-intensive versions align more closely with PEFT methods rather than fine-tuning-free low-rank approximation approaches. For instance, when applied at the model level, ApiQ effectively becomes equivalent to training LoRA on top of a compressed model—shifting its focus toward fine-tuning rather than fine-tuning-free compensation, and thus falling outside the scope of this study. Note that the optimization time for both \methodabbr~and Act-S is comparable, with each completing within minutes, whereas ApiQ requires over hours to optimize.

We evaluate EoRA and the baselines on improving the task-specific accuracy of the compressed LLMs on language generation, commonsense reasoning, and math reasoning tasks using the LM-Evaluation-Harness framework~\citep{eval-harness}. We pick WikiText2 for the language generation task and perplexity as the evaluation metric. For commonsense reasoning, we select ARC-Challenge (ARC-C) \citep{clark2018think}, and for math reasoning ability, we choose MathQA \citep{amini2019mathqa} and GSM8K \citep{cobbe2021training}. We sample 128 concatenated sentences of length 2048 from the WikiText2 training set as the calibration set for \methodabbr, Act-S, and ApiQ for the language generation task. For commonsense reasoning tasks, we sample 32 concatenated sentences of length 2048 from the ARC training set and combine them with 32 concatenated sentences of the same length from C4~\citep{raffel2020c4} to construct the calibration set for \methodabbr, Act-S, and ApiQ. Similarly, for the math reasoning tasks, we sample 32 concatenated sentences of length 2048 from the MathQA/GSM8K training set and combine them with 32 concatenated sentences from C4 to form the calibration set for the three methods.

\subsection{Main Results}
\subsubsection{Sparsity Error Compensation}
\input{tables/main_table_sparsegpt_r128}
To assess the effectiveness of \methodabbr~in compensating for sparsity error, we compare \methodabbr~with all the baselines on LLaMA2-7B/13B and LLaMA3-8B models pruned with SparseGPT to 2:4 sparsity—the only sparsity format that yields actual inference speedups on GPUs. Rank of all the compensation methods is set to 128, and the results of LLaMA3-8B are presented in Table~\ref{tab:sparsegpt_r128}, while the full results, including LLaMA2-7B/13B, are provided in Table~\ref{tab:appendix_sparsegpt_r128} in the appendix. \methodabbr~consistently outperforms all fine-tuning-free baselines, achieving gains of $2.9\%$, $2.1\%$, and $10.7\%$ over Act-S on ARC-C, MathQA, and GSM8K, respectively. Furthermore, it surpasses ApiQ by $0.4\%$ on ARC-C and $1.1\%$ on MathQA, while delivering comparable results on GSM8K—all with significantly faster optimization time (\textit{15 minutes vs. 2.5 hours}). Furthermore, \methodabbr~proves robustness across different model sizes, continuing to outperform ZeroQuant-V2, Act-S, and ApiQ in boosting the accuracy of 2:4 pruned LLaMA2-7B/13B across ARC-C and MathQA as shown in Table~\ref{tab:appendix_sparsegpt_r128}. We further assess the generalizability and compatibility of EoRA with pruning methods beyond SparseGPT. Specifically, we evaluate \methodabbr~on LLaMA3-8B pruned to 2:4 sparsity using Wanda~\citep{sun2023simple}, where \methodabbr~continues to outperform all the fine-tuning-free baseline methods. For additional details, please refer to Section~\ref{sec:other_compression_methods}.

\subsubsection{Quantization Error Compensation}
\input{tables/main_table_gptq_r128}
We evaluate \methodabbr~on LLaMA2-7B/13B and LLaMA3-8B models quantized with GPTQ to 4-bit and 3-bit to assess the effectiveness of \methodabbr~in compensating for quantization error. The ranks for all the methods are set to 128. From Table~\ref{tab:gptq_r128}, 3-bit quantization causes significant accuracy degradation, with losses up to $29.5\%$/$17.7\%$/$35.8\%$ on ARC-C, MathQA, and GSM8K, respectively. By applying \methodabbr, we demonstrate that the accuracy loss can be reduced to $18.7\%$/$10.9\%$/$24.3\%$ on ARC-C, MathQA, and GSM8K—providing $10.8\%$/$6.7\%$/$11.5\%$ improvement, outperforming all the baseline methods for compensating the quantization error. On the other hand, although 4-bit quantization does not result in as much accuracy loss as 3-bit quantization, applying \methodabbr~can still generally enhance the performance of the 4-bit model, offering up to a $2.2\%$ and $3.14\%$ accuracy boost on ARC-C and MathQA, respectively. Comprehensive results, including those for LLaMA2-7B/13B, are presented in Table~\ref{tab:appendix_gptq_r128} in the appendix, where a similar trend of improvement with \methodabbr~is observed. We also explore the feasibility of using \methodabbr~to improve ultra-compressed models that combine both pruning and quantization. In this setting, \methodabbr~continues to outperform all baselines on ARC-C and MathQA. Further details can be found in Section~\ref{sec:sparse_and_quant} of the Appendix.

\subsection{Ablation Study: Ranks and Calibration Sizes}
\begin{figure}[ht]
\begin{center}
\includegraphics[width=1.0\columnwidth]{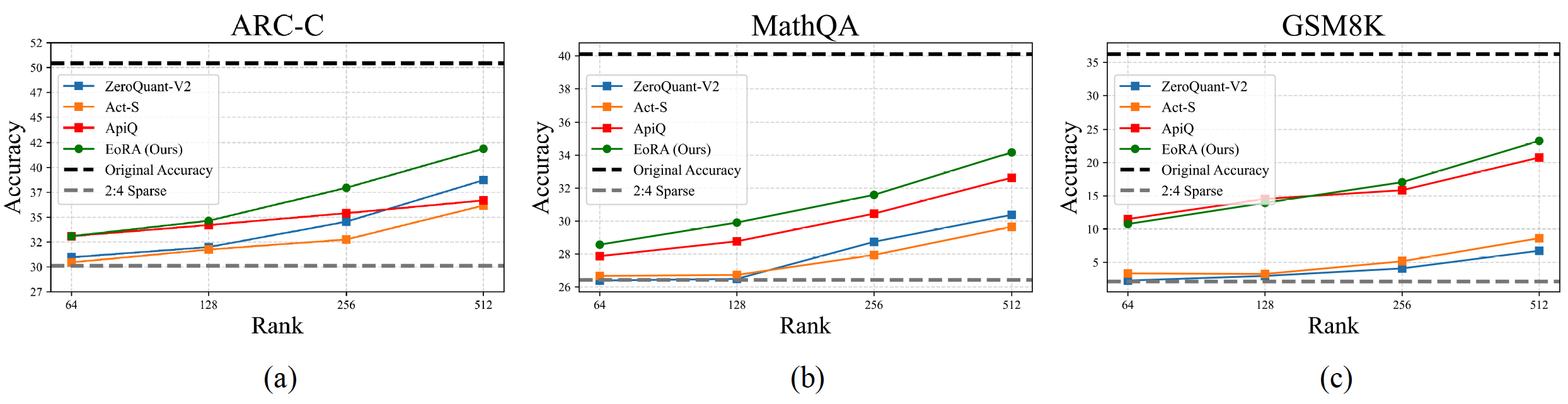}
\end{center}
\caption{Results of applying~\methodabbr and other baselines with rank set to \{64,128,256,512\} to improve LLaMA3-8B models pruned to 2:4 sparsity by SparseGPT on (a) ARC-C/(b) MathQA/(c) GSM8K.}
\label{fig:diff_rank}
\end{figure}
\noindent
Since one of the advantages of using \methodabbr~is the greater flexibility in adjusting overall model accuracy without being constrained by specific compression formats, in this section, we investigate the influence of different ranks on adopting \methodabbr. We vary the rank of \methodabbr~in \{64,128,256,512\} on compensating LLaMA3-8B pruned to 2:4 sparsity. As shown in Figure~\ref{fig:diff_rank}, \methodabbr~consistently outperforms the two fine-tuning-free baselines (ZeroQuant-V2 and Act-S) across all tested ranks, with the performance gap becoming more prominent at higher ranks. For instance, on GSM8K, \methodabbr~achieves improvements of $7.43\%$, $10.69\%$, $11.9\%$, and $14.62\%$ at ranks 64, 128, 256, and 512, respectively. In contrast, the gains on ARC-C remain relatively stable across ranks, ranging between $2\%$ and $4\%$. Additionally, \methodabbr~begins to outperform ApiQ on GSM8K at higher ranks, with improvements of $1.21\%$ and $2.51\%$ observed at ranks 256 and 512, respectively. These experiments prove that \methodabbr~is robust across different rank settings, offering users a more flexible option upon existing compression configurations to effectively balance the trade-off between inference overhead and model accuracy. A similar trend is observed in the results for LLaMA2-7B/13B shown in Table~\ref{tab:appendix_diff_rank} in the appendix.

We also compare the influence of different calibration sizes on \methodabbr~. We vary the calibration size in \{16,32,64\}, and compare them on recovering the accuracy of LLaMA3-8B quantized to 3/4-bit and pruned to 2:4 sparsity. Overall, we find that \methodabbr~demonstrates strong robustness and maintains competitive accuracy even with limited calibration data, as shown in Table~\ref{tab:appendix_calib}. Notably, using as few as 32 calibration samples to compensate for a 2:4 pruned model can even yield better accuracy improvements than using 64 samples.

\subsection{\methodabbr~as LoRA initialization for Fine-tuning Compressed Models}
\label{sec:finetuning_trex}
\input{tables/finetuning_table}
\noindent
We show that, with additional computational resources, users can leverage the low-rank matrices from \methodabbr~as initialization for LoRA fine-tuning, enabling further accuracy improvements for compressed models. We follow the conventional LoRA fine-tuning framework, which keeps the compressed model frozen and only tunes the low-rank residual components during fine-tuning. We conduct experiments on compressed LLaMA3-8B models with \{2:4 sparsity, 4-bit, 3-bit\} compression. The rank of LoRA is set to 128 and is applied to every linear layer, initialized using \methodabbr, SVD following LoftQ~\citep{liloftq}, and standard initialization following QLoRA~\citep{dettmers2024qlora}. Fine-tuning is performed on the ARC training set for evaluating ARC-C, and on the MathQA training set for the math reasoning task. We fine-tune the models for 3 epochs with a batch size of 64, a learning rate of 1e-5, and a cosine learning rate scheduler. As shown in Table~\ref{tab:finetuning}, initializing with \methodabbr~substantially enhances the accuracy of compressed models, surpassing both QLoRA and LoftQ when fine-tuning 4-bit quantized LLaMA3-8B, and achieving accuracy on par with standard full-precision fine-tuning. We also observed that the improvements over QLoRA and LoftQ are more pronounced on 3-bit quantized and 2:4 pruned models, aligning with our earlier finding that \methodabbr~is more effective when the compression error is more substantial, as shown in Table~\ref{tab:appendix_finetuning}.

\subsection{Kernel Optimization, Inference Speed Evaluation and Memory Overhead of EoRA}
\label{model-size}
\begin{figure}[ht]
\begin{center}
\includegraphics[width=1.0\columnwidth]{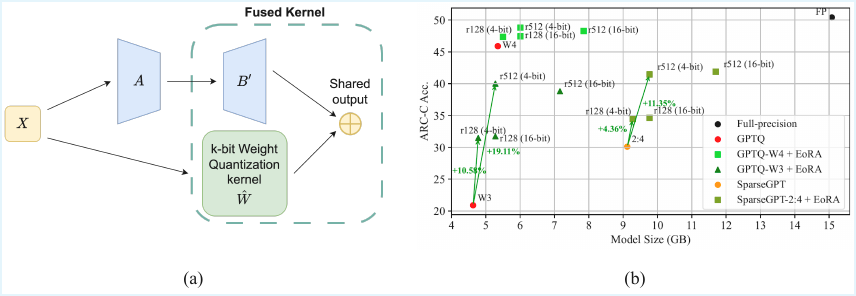}
\end{center}
\caption{(a) We propose fusing the multiplication of $B$ with the weight quantization kernel to minimize data movement overhead and substantially improve the inference latency. (b) The model size and ARC-C accuracy of EoRA with rank 128/512, quantized to 4-bit for compensating LLaMA3-8B quantized to 4/3-bit or pruned to 2:4 sparsity.}
\label{fused-kernel}
\end{figure}

While theoretically, compensating a compressed model with low-rank residual paths introduces minimal computational overhead, in practice, it leads to a noticeable increase in latency. This is primarily because input and output must transfer between L2 cache and DRAM twice as often compared to that without a low-rank residual path, shifting the inference process from being computation-bound to memory-bound. This phenomenon is also discussed in ~\citep{li2024svdqunat}. To address this, we propose fusing the low-bit weight quantization kernel with the matrix multiplication of $B$, which shares the same output. By doing so, the shared output no longer needs to be offloaded and reloaded to the L2 cache, effectively reducing data transfer overhead as illustrated in Figure~\ref{fused-kernel} (a). Implementation details of our kernel can be found in Section~\ref{appendix_kernel}. As shown in Table \ref{tab:speedup}, our custom EoRA kernel substantially accelerates inference compared to using native PyTorch for the low-rank residual path on top of the low-bit quantized kernel, achieving a speedup of up to 1.4x over FP16 with EoRA of rank 128 at 3-bit quantization. In contrast, without the EoRA kernel, the initial 1.7x speedup provided by the 3-bit quantized kernel drops to 1.1x. Similarly, under 4-bit quantization, the EoRA kernel delivers an extra 0.3x speedup compared to setups without the EoRA kernel.

Finally, \methodabbr~can also be quantized to further reduce the additional cost of residual low-rank compensation paths. In this section, we quantize \methodabbr~of rank \{128, 512\} to 4/3-bit on compensating three types of compressed LLaMA3-8B models (2:4 pruned, 4-bit quantized, and 3-bit quantized). The complete results are provided in Table~\ref{tab:appendix_quant_eora} in appendix, while the results for LLaMA3-8B are illustrated in Figure~\ref{fused-kernel} (b). As shown in the figure, \methodabbr~is robust to quantization, which means that when \methodabbr~is quantized, the accuracy drop from full-precision \methodabbr~is insignificant while the model size is significantly reduced. For example, when a 512-rank \methodabbr~is quantized from 16-bits to 4-bit on 2:4 pruned LLaMA3-8B, the accuracy drops are only $0.43\%$ on ARC-C while the total model size reduces by $16.49\%$. Additionally, compared to the original uncompensated 2:4 pruned model, quantizing \methodabbr~of rank 128/512 improves accuracy by $4.4\%$/$11.4\%$ with a total model size increase of just $2\%$/$7\%$. For 3-bit quantized LLaMA3-8B compensated with a 4-bit quantized \methodabbr~of rank 128/512 achieves $10.6\%$/$19.1\%$ accuracy improvements, with a corresponding model size increase of only $3\%$/$14\%$. Interestingly, we also observe that quantizing \methodabbr~does not always result in accuracy loss; in some cases, it even slightly improves accuracy, potentially due to quantization acting as a form of regularization, as discussed in OFQ~\citep{liu2023oscillation}. Generally, we recommend users quantize \methodabbr~to 4-bit, as this significantly reduces inference latency and model size with kernel support, without causing any noticeable drop in accuracy.

\section{Related Works}

\paragraph{Post-training LLM Compression:} As LLMs scale, reducing their size is essential for efficient deployment. Traditional compression-aware training methods are impractical due to the need for full datasets and heavy retraining. Post-training compression methods like quantization and pruning have gained popularity as they require only minimal calibration data and no retraining. PTQ reduces model size by lowering bitwidths \citep{frantar2022gptq, tseng2024quip}, while PTP removes less important weights to reduce computation \citep{frantar2023sparsegpt, sun2023simple}. Our method, \methodabbr, is compatible with all such compression techniques as it operates independently of the base method used.

\paragraph{Low-Rank Decomposition:} Low-rank decomposition methods \citep{yuan2023asvd, sakr2024espace, wang2024svdllm, hsu2022fwsvd, mozaffari2024slim, zhang2024oats, zhang2024lqer, zhang2024qera, saha2024compressing} compress models by replacing weights with low-rank matrices, reducing both latency and size without special kernel support. However, they are less widely adopted due to weaker accuracy-compression trade-offs. 
While FWSVD~\citep{hsu2022fwsvd} is designed primarily as a compression method based on low-rank decomposition of model weights, \methodabbr~instead leverages low-rank modules specifically to compensate for compression errors.
Unlike FWSVD, \methodabbr~provides a theoretical guarantee of minimizing layer-wise compression loss, as demonstrated in our derivation in Section~\ref{sec:method}. Using gradient-based information, as required by FWSVD, can be prohibitively expensive for LLMs, as noted in ASVD~\citep{yuan2023asvd}. 
SVD-LLM \citep{wang2024svdllm} is conceptually close to our work, which also tries to align the SVD compression error with the layer-wise compression loss. It relies on the matrix product of the activation being positive-definite, a condition often unmet in practice. Enforcing this condition typically requires additional modifications, which introduce noise into the approximation. In contrast, \methodabbr~employs eigendecomposition, which only requires the matrix product of the activation to be symmetric—a property that naturally holds—avoiding such issues. Furthermore, although \methodabbr~also utilizes SVD-based low-rank decomposition, its core objective is fundamentally different. Whereas prior methods aim to replace pre-trained weight matrices with low-rank approximations to reduce model size and inference cost, \methodabbr~focuses on approximating the compression error itself. This allows for improved accuracy recovery in compressed LLMs and provides greater flexibility in balancing accuracy and computational overhead by overcoming the constraints of fixed compression formats. Please refer to Section~\ref{sec:recent_comparison} in Appendix for more detailed comparison.

\section{Conclusion}
In this work, we present \methodabbr, a novel fine-tuning-free approach that rapidly boosts the task-specific accuracy of compressed LLMs using minimal calibration data, while offering greater flexibility by relaxing compression format constraints. By projecting compression errors into the task-specific eigenspace of activations, \methodabbr~uses eigenvalues to guide SVD, aligning approximation error with layer-wise compression loss—without any gradient-based training. \methodabbr~achieves strong results across language, commonsense, and math reasoning tasks, outperforming prior low-rank methods~\citep{yao2024zeroquantv2, yuan2023asvd, liao2024apiq}. Its fine-tuning-free design allows quick adaptation to various accuracy-latency trade-offs, and it remains robust under quantization, reducing memory overhead. Additionally, it can serve as a strong initialization for LoRA fine-tuning. Overall, \methodabbr~is a scalable, efficient solution for improving compressed LLMs across diverse deployment settings, with potential extensions to new architectures and modalities.

{
\small
\bibliography{citation}
\bibliographystyle{unsrt}
}

\newpage
\appendix
\section{Appendix}
\subsection{Sparsity Error Compensation}
\input{tables/appendix_sparse_table}
Table~\ref{tab:appendix_sparsegpt_r128} reports more detailed comparison for sparsity error compensation.

\subsection{Quantization Error Compensation}
\input{tables/appendix_quant_table}
Table~\ref{tab:appendix_gptq_r128} reports more detailed comparison for quantization error compensation.
It is worth noting that EoRA-enhanced models outperform smaller models quantized at higher precision, while high-precision models usually outperform low-precision models. For example on GSM8K (Table~\ref{tab:appendix_gptq_r128}), 3-bit LLaMA2-13B (4.62) performs worse than 4-bit LLaMA2-7B (9.93), but EoRA-enhanced 3-bit LLaMA2-13B (15.08) outperforms 4-bit LLaMA2-7B. 
This highlights EoRA’s advantage in terms of greater flexibility: rather than shrinking the model architecture, it enables more effective compression of larger models while preserving higher accuracy.
\newpage

\subsection{Sparsity \& Quantization Error Compensation}
\input{tables/appendix_sparse_and_quant_table}
\label{sec:sparse_and_quant}
Here, we examine the feasibility of applying \methodabbr~to compensate for ultra-compressed models that undergo both pruning and quantization, as shown in Table~\ref{tab:appendix_sparse_and_quant}. Specifically, we prune LLaMA2-7B/13B and LLaMA3-8B to 2:4 sparsity and quantize them to 4-bit. We set the ranks of both \methodabbr~and SVD to 128 to compensate for the pruning and quantization errors. Similarly to our previous findings, LLaMA3-8B is the least resilient to compression, experiencing a significant drop in both perplexity for language generation and accuracy on commonsense and math reasoning tasks. Notably, the accuracy on ARC-C plummets to $18.33\%$ and MathQA to $19.89\%,$ which is worse than random guessing. However, compensating for the sparsity and quantization errors with \methodabbr~significantly improves the accuracy of these compressed models, reducing perplexity by up to $73.55$ and boosting accuracy by $12.88\%$/$9.60\%$/$10.16\%$ on ARC-C/MathQA/GSM8K tasks. Additionally, \methodabbr~consistently outperforms both ZeroQuant-V2 and Act-S across LLaMA2 and LLaMA3. For instance, \methodabbr~exceeds ZeroQuant-V2 in compensating the compressed LLaMA2-13B on ARC-C by $1.79\%$ and on MathQA by $1.77\%,$ narrowing the accuracy gap with the uncompressed model to just $2.85\%$ on MathQA. Overall, we find that \methodabbr~tends to offer greater accuracy recovery when addressing more aggressive compression settings, ensuring the plausibility of adopting \methodabbr~for mitigating severe compression error. 
\newpage

\subsection{Compatibility With Various Compression Methods}
\label{sec:other_compression_methods}
\input{tables/different_compression}
In this section, we study the generalizability and compatibility of \methodabbr~with different pruning methods beyond SparseGPT. We adopt Wanda~\citep{sun2023simple}, a method that prunes weights with the smallest magnitudes scaled by their corresponding input activations. For these compression methods, we adhere to the calibration set construction detailed in \ref{sec:experiments_details}, and maintain the same settings when utilizing \methodabbr~to address compression errors. We evaluate \methodabbr on LLaMA3-8B pruned with Wanda to 2:4 structured sparsity. The ranks of all low-rank compensation methods are set to 128. Table~\ref{tab:dif_compression} demonstrates that \methodabbr consistently outperforms every fine-tuning-free method, both ZeroQuant-V2 and Act-S, in improving accuracy across all the tasks. For example, \methodabbr~achieves accuracy gains of 7.77\%/4.96\%/10.76\% on ARC-C/MathQA/GSM8K which is  5.04\%/3.32\%/10.01\% over the improvement brought by Act-s. Furthermore, EoRA outperforms ApiQ on both ARC-C and MathQA by 2.9\% and 0.44\%. Overall, these findings underscore the effectiveness and generalizability of EoRA across different compression techniques.

\subsection{Influence of Different Calibration sizes on \methodabbr}
\input{tables/appendix_calib}

We conducted ablation studies to assess how different calibration set sizes—\{16, 32, 64, 128, 256, 512\}—affect EoRA’s performance in compensating for 4-bit and 3-bit quantized LLaMA3-8B models on MathQA. As shown in Table~\ref{tab:appendix_calib}, increasing the number of calibration samples from 16 to 256 yields a moderate improvement in accuracy (from 36.62 to 37.60) for the W4 model. However, further increasing the calibration size to 512 leads to a slight decline, indicating that the accuracy gain saturates beyond a certain threshold. A similar trend is observed for the W3 model, where performance improves steadily up to 128 samples, after which the benefit plateaus. These findings suggest that while EoRA can leverage additional calibration data to improve accuracy, its performance remains stable and robust even with limited calibration.

\subsection{Inference Speed Evaluation}
\label{appendix_kernel}
\input{tables/speedup}
In language generation, the model produces tokens sequentially, making matrix-vector multiplications the primary factor impacting the inference latency. Consequently, we build our custom EoRA kernel on top of GPTQ's low-bit quantized matrix vector product kernel, pre-allocating the shared output prior to matrix vector multiplication and integrating the full-precision matrix vector multiplication of $B$ into the quantized kernel reducing redundant memory access. We show the inference speedup of our proposed EoRA kernel in Table~\ref{tab:speedup}. 
The first row shows the FP16 latency (60ms), followed by the 3-bit quantized-only model (35ms). The remaining rows present EoRA latencies at different ranks, both with and without our custom CUDA kernel, alongside 3-bit and 4-bit quantization. 
As shown in the table, our fused EoRA kernel significantly improves inference speed for both 3-bit and 4-bit quantized models. 
While there remains some overhead compared to the quantized-only baseline, even with our kernel, using no kernel results in significantly higher latency—sometimes exceeding FP16—due to activation movement overhead, as discussed in Section~\ref{model-size}.
These results underscore the practical deployability of EoRA when paired with our optimized kernel.
\newpage

\subsection{Compensation With Different Ranks}
\input{tables/appendix_different_rank}
Table~\ref{tab:appendix_diff_rank} reports more detailed comparison for error compensation with different ranks.
\newpage

\subsection{Fine-tuning Compressed Models with EoRA}
\label{appendix_finetuning}
\input{tables/appendix_finetuning}
Table~\ref{tab:appendix_finetuning} reports more detailed comparison for fine-tuning compressed models.

\subsubsection{Ablation: Fine-tuning with different numbers of training data}
\input{tables/finetuning_diff_dataset_ratio_table}
\noindent
In this section, we show that fine-tuning with the \methodabbr-compensated model is robust to various ratios of training data.
We follow the setting in Sec.~\ref{sec:finetuning_trex}
on compressed LLaMA3-8B models with 2:4 sparsity compression. As shown in Table~\ref{tab:finetuning_diff_dataset_ratio}, using \methodabbr~for initialization consistently outperforms both standard and SVD initialization across various dataset ratios, with accuracy improvements (ARC-E/ARC-C/MathQA) of $3.24\%$/$4.95\%$/$6.4\%$ and $1.85\%$/$4.09\%$/$4.19\%$ over SVD when fine-tuning using $50\%$ and $30\%$ training data, respectively. 

\subsection{Quantizing EoRA To Further Reduce Memory Overhead}
\input{tables/appendix_lowrank_quantization}
\label{sec:size_eora}
Table~\ref{tab:appendix_quant_eora} reports more detailed comparison for quantizing EoRA.
For example, when a 512-rank \methodabbr~is quantized from 16-bits to 4-bit on 2:4 pruned LLaMA3-8B, the accuracy drops are only $0.43\%$ on ARC-C while the total model size reduces by $16.49\%$ (11.70 GB $\rightarrow$ 9.77 GB). Additionally, compared to the original uncompensated 2:4 pruned model, quantizing \methodabbr~of rank 128/512 improves accuracy by $4.4\%$/$11.4\%$ with a total model size increase of just $2\%$ (9.12 GB $\rightarrow$ 9.28 GB) / $7\%$ (9.12 GB $\rightarrow$ 9.77 GB). For 3-bit quantized LLaMA3-8B compensated with a 4-bit quantized \methodabbr~of rank 128/512 achieves $10.6\%$/$19.1\%$ accuracy improvements, with a corresponding model size increase of only $3\%$ (4.63 GB $\rightarrow$ 4.78 GB) / $14\%$ (4.63 GB $\rightarrow$ 5.28 GB). Please see Section~\ref{model-size} for more analyses. 
\newpage

\subsection{EoRA on More Tasks}
\input{tables/llm_summarization}
\paragraph{LLM Summarization.}
We evaluate EoRA and baseline methods on restoring the summarization capability of quantized models using the testset of CNN/DailyMail — a widely used English-language corpus containing over 300k news articles authored by CNN and Daily Mail journalists. This dataset supports both extractive and abstractive summarization, and we adopt ROUGE-Lsum as the evaluation metric, where higher values indicate better summary quality. We set the rank to 128 for all the methods. For Act-S, ApiQ, and EoRA, we use 128 calibration sentences from WikiText2. The results are shown in Table~\ref{tab:llm_summarization}. Notably, EoRA consistently outperforms all baselines across both 4-bit and 3-bit quantized LLaMA3-8B models. Specifically, it improves the ROUGE-Lsum score from 0.1672 to 0.1812 in the 4-bit setting, and from 0.0650 to 0.1463 in the 3-bit setting—demonstrating substantial recovery of summarization performance. These results highlight that EoRA remains effective and competitive in practical, real-world scenarios such as summarization. 

\input{tables/mmlu}

\paragraph{Language Understanding.}
We tested EoRA on 4-bit quantized LLaMA-3.2-3B, varying the rank from 8 to 128. Using 128 WikiText2 calibration samples, we observe substantial recovery—even rank 8 lifts accuracy from 24.16 to 52.20, as shown in Table~\ref{tab:mmlu}. This result highlights both the effectiveness of EoRA in compensating for quantization errors in smaller-scale models and its robustness across a range of low-rank settings.
\newpage

\subsection{GPTQ: Channel-wise vs. Group-wise Quantization}
\input{tables/groupwise_eora}

We initially adopt channel-wise quantization for GPTQ to remain consistent with the original GPTQ setup. More importantly, our goal is to showcase the robustness of EoRA in recovering from even severe quantization errors. That said, we conduct additional experiments using group-wise quantization (group size = 128) to evaluate EoRA's effectiveness in this more commonly used setting. 
Table~\ref{tab:groupwise} reports results on MathQA for 4-bit and 3-bit group-wise quantized LLaMA3-8B models, referred to as W4-groupsize 128 and W3-groupsize 128, respectively. We compare EoRA against prior compensation methods and set the rank to 128.
These results highlight EoRA’s consistent advantage over existing methods in compensating for group-wise quantization. In particular, for the W4-group size 128 configuration, EoRA is able to recover nearly all of the lost accuracy, achieving performance comparable to the original full-precision model. 

\subsection{Layer-wise Discrepancy Analysis}
\input{tables/layerwise_discrepancy}
We follow the layer-wise discrepancy analysis setting of the ApiQ~\citep{liao2024apiq} and run the analysis on the $q$ projector of LLaMA2-7B. We compare the discrepancy of layers [0, 5, 10, 15, 20, 25, 30] of different compensation methods of rank 128 and show that EoRA effectively reduces layer-wise discrepancy compared to existing baselines, as shown in Table~\ref{tab:layerwise_discrepancy}.
Notably, EoRA maintains consistently low output activation errors across all layers, especially in the later ones (e.g., 5.8 at layer 20 and 5.4 at layer 30), showing significant improvements over ZeroQuant-V2 and Act-S. EoRA also matches or outperforms ApiQ, demonstrating its effectiveness. Importantly, EoRA accomplishes this with dramatically less compute overhead—typically completing in just a few minutes—whereas ApiQ often requires several hours of optimization. This highlights EoRA’s practical advantage as a highly efficient solution for layer-wise error minimization for recovering the error of low-bit quantization.

\subsection{More Comparisons with Recent Low-Rank Approaches}
\label{sec:recent_comparison}
\input{tables/more_comparison}

Recent Low-Rank approaches can be broadly categorized into three groups: 1) activation-statistics-based scaling methods, including SLiM~\citep{mozaffari2024slim}, OATS~\citep{zhang2024oats}, LQER~\citep{zhang2024lqer}, and QERA~\citep{zhang2024qera}, 2) iterative low-rank compensation approaches, which are LRC~\citep{scetbon2024low} and CALDERA~\citep{saha2024compressing}, and 3) training-based methods, including LR-QAT~\citep{bondarenko2024low} and SLoPe~\citep{mozaffari2024slope}. 

The first group—SLiM, OATS, LQER, and QERA—scales the compression error based on activation statistics prior to applying SVD for low-rank approximation. The underlying intuition is that input channels with higher magnitudes (i.e., outliers) are considered more important, and as a result, their corresponding weight entries should be prioritized over those linked to lower-magnitude activations. All four methods differ slightly in how the scaling diagonal matrix is constructed, but they are fundamentally similar to the ASVD-based scaling baseline already discussed in our paper (see Section~\ref{sec:experiments_details}). 
For example, LQER~\citep{zhang2024lqer} mitigates quantization error by scaling the residual using activation statistics—specifically, the maximum average magnitude per input channel—before applying SVD. 
Another example, SLiM~\citep{mozaffari2024slim}, introduces a low-rank compensation method that also leverages activation statistics (specifically, average absolute activation values) to scale residuals. Although SLIM is proposed alongside a custom quantization scheme, we isolate and evaluate its saliency-based low-rank adapter strategy. 
Although these methods vary in how they scale the compression error, none of them are guaranteed to minimize the layer-wise compression error directly—an issue we discussed in Section~\ref{sec:preliminaries} of our paper. 
In contrast, EoRA is explicitly formulated to minimize this objective (as detailed in Section~\ref{sec:method}). 

The second group, including iterative low-rank compensation approaches, mainly lack the flexibility as EoRA. For example, LRC~\citep{scetbon2024low} requires iterative updates to weights and low-rank modules, leading to task-specific quantized models. In contrast, EoRA only adapts the low-rank modules, allowing a shared compressed backbone and easier integration with multi-adapter frameworks~\citep{multilora}. Although LRC offers a closed-form solution, it assumes is full-rank—an assumption that often fails and requires extra modification steps that may introduce noise and instability. EoRA avoids this by only requiring to be symmetric, improving numerical stability and robustness. 
Another example, CALDERA~\citep{saha2024compressing}, employs an iterative optimization strategy that updates both the quantized weights and the low-rank matrices, using a closed-form solution. However, because it requires modifying the quantized weights during this process, CALDERA is less efficient for multi-task scenarios, where separate quantized models would need to be maintained for each task. EoRA, on the other hand, avoids this limitation by making only the low-rank components task-specific, while keeping the quantized backbone fixed and shared across tasks. This decoupled design allows for easy integration with existing multi-adapter inference frameworks~\citep{multilora} and significantly improves the practicality of EoRA for real-world applications. 

The third group requires gradient-based training. For example, LR-QAT~\citep{bondarenko2024low} is a quantization-aware training (QAT) method. While it also uses low-rank modules, it operates in the fine-tuning regime, combining cross-entropy loss with final-layer output alignment (akin to knowledge distillation). SLoPe~\citep{mozaffari2024slope} primarily targets improving pre-training efficiency and enhancing the accuracy of compressed models through training. As such, it aligns more closely with quantization-aware training (QAT) as well.
In contrast, EoRA is a post-training quantization (PTQ) method that requires no gradient updates to the quantized model. In the LLM compression community~\citep{frantar2022gptq,lin2024awq}, it is standard practice to distinguish between PTQ and QAT approaches, as they serve different purposes and are not typically benchmarked against each other.

We summarize the empirical comparison on 4-bit and 3-bit quantized LLaMA3-8B models evaluated on MathQA in Table~\ref{tab:more_comparison}. All methods use rank of 128 and identical calibration data (see Section~\ref{sec:experiments_details}). To fairly compare CALDERA with EoRA, we adapt the CALDERA method by fixing the quantized weights and only updating the low-rank matrices. This degenerates the iterative process into a two-step approximation: first approximate the down-projection matrix, followed by the up-projection. Once the quantized weights are fixed, additional iterations do not change the approximation. 

As the results show, EoRA consistently outperforms other low-rank compensation methods, particularly those based on activation statistics like LQER, SLiM, etc. This highlights the advantage of EoRA’s mathematical property, which directly minimizes the layer-wise compression error rather than relying on heuristics derived from activation statistics.
While LRC offers better accuracy than heuristic scaling approaches, it still lags behind EoRA due to its reliance on less stable approximations. These findings highlight EoRA’s practicality and effectiveness for post-training quantized LLMs. 
From the results, we observe that CALDERA performs well—outperforming both Act-S and ApiQ—but EoRA still consistently achieves higher accuracy. We attribute this to EoRA’s single-step optimization that jointly solves for both projection matrices, whereas CALDERA’s sequential two-step process may accumulate slightly more approximation error.

Overall, EoRA consistently surpasses all the recent low-rank methods, including both activation-statistics-based approaches and iterative low-rank compensation approaches. Its effectiveness stems from its ability to directly minimize layer-wise compression error, eliminate the need for heuristic magnitude-based scaling, and utilize a single-step optimization process. These strengths underscore both the theoretical robustness and practical efficiency of EoRA compared to existing methods.

\subsection{EoRA on 2-bit Quantization}
\input{tables/2bit_gptq}

We also evaluate a more challenging setting, 2-bit quantization, where RILQ~\citep{lee2025rilq} is one of the state-of-the-art error compensation methods. RILQ adopts standard backpropagation (i.e., fine-tuning), utilizing a combination of cross-entropy loss and final-layer output alignment (similar to knowledge distillation). As outlined in the RILQ paper, RILQ uses gradient descent to collectively tune all adapters, minimizing the discrepancy between full-precision and quantized activation outputs of the final layer. In addition, as quoted from the RILQ paper, RILQ also incorporates a causal language modeling objective with Ground Truth in the optimization of low-rank adapters. Therefore, it is not appropriate to directly compare EoRA to RILQ as methods in the same category.
However, since EoRA can act as an effective initialization for subsequent fine-tuning (as detailed in Section~\ref{sec:finetuning_trex}), it is complementary to RILQ rather than competing with it. To examine this synergy, we performed experiments where EoRA was first used for initialization—following the setup outlined in Section~\ref{sec:finetuning_trex}—and then applied RILQ’s fine-tuning objective to a 2-bit GPTQ-quantized LLaMA3-8B model on the MathQA dataset. The results are shown in Table~\ref{tab:2bit_gptq}. As expected, RILQ outperforms LoftQ in compensating for quantization error in 2-bit models. Moreover, when initialized with EoRA, further fine-tuning with RILQ’s objective yields an additional 1.3\% improvement over RILQ alone.

%% file: tables/main_table_sparsegpt_r128.tex
\begin{table*}[!htp]\centering
\caption{Perplexity and commonsense/math reasoning results for LLaMA3-8B pruned with 2:4 sparsity using SparseGPT, with all compensation methods evaluated at rank 128.}\label{tab:sparsegpt_r128}
\resizebox{1\columnwidth}{!}{
\begin{tabular}{cccccccc}\toprule
Model &Sparsity &Compensation Method &Wikitext2 $\downarrow$ &ARC-C $\uparrow$&MathQA $\uparrow$&GSM8K $\uparrow$\\\cmidrule{1-7}
\multirow{8}{*}{LLaMA3-8B} &- &- &6.13 &50.42 &40.10 &36.23 \\\cmidrule{2-7}
&\multirow{7}{*}{2:4} &- &12.32 &30.11 &26.43 &2.12 \\\cmidrule{3-7}
& &ZeroQuant-V2 &11.31 &31.99 &26.49 &2.956 \\\cmidrule{3-7}
& &Act-S &11.32 &31.74 &26.73 &3.26 \\\cmidrule{3-7}
& &ApiQ &\underline{11.08} &\underline{34.21} &\underline{28.77} &\textbf{14.55} \\\cmidrule{3-7}
& &EoRA (Ours) &\textbf{11.07} &\textbf{34.64} &\textbf{29.91} &\underline{13.95} \\\cmidrule{1-7}
\bottomrule
\end{tabular}}
\end{table*}

%% file: tables/main_table_gptq_r128.tex
\begin{table}[!htp]\centering
\caption{Perplexity and commonsense/math reasoning results for LLaMA3-8B quantized to 3/4-bits using GPTQ, with all compensation methods evaluated at rank 128.}\label{tab:gptq_r128}
\resizebox{1\columnwidth}{!}{
\begin{tabular}{cccccccc}\toprule
Model &W bits &Compensation Method &Wikitext2 $\downarrow$ &ARC-C $\uparrow$&MathQA $\uparrow$&GSM8K $\uparrow$ \\\cmidrule{1-7}
\multirow{16}{*}{LLaMA3-8B} &- &- &6.13 &50.42 &40.10 &36.23 \\\cmidrule{2-7}
&\multirow{8}{*}{W4} &- &7.00 &45.90 &34.07 &27.74 \\\cmidrule{3-7}
& &ZeroQuant-V2 &\underline{6.80} &45.24 &\underline{36.51} &\textbf{31.23} \\\cmidrule{3-7}
& &Act-S &6.82 &\textbf{47.86} &35.84 &29.34 \\\cmidrule{3-7}
& &ApiQ &6.87 &46.58 &36.18 &30.09 \\\cmidrule{3-7}
& &EoRA (Ours) &\textbf{6.80} &\underline{47.44} &\textbf{37.21} &\underline{30.70} \\\cmidrule{2-7}
&\multirow{8}{*}{W3} &- &15.64 &20.90 &22.37 &0.45 \\\cmidrule{3-7}
& &ZeroQuant-V2 &10.24 &30.02 &26.43 &3.79 \\\cmidrule{3-7}
& &Act-S &\underline{10.19} &\underline{31.28} &25.42 &4.09 \\\cmidrule{3-7}
& &ApiQ &10.41 &30.46 &\underline{26.86} &\underline{10.79} \\\cmidrule{3-7}
& &EoRA (Ours) &\textbf{10.06} &\textbf{31.74} &\textbf{29.11} &\textbf{11.90} \\\cmidrule{1-7}
\bottomrule
\end{tabular}}
\end{table}

%% file: tables/finetuning_table.tex
\begin{table}[!htp]\centering
\caption{Finetune the 4-bit compressed LLaMA3-8B models with different initialization of the low-rank matrices for Commonsense/Math reasoning tasks.}\label{tab:finetuning}
\resizebox{1\columnwidth}{!}{
\begin{tabular}{ccccccc}\toprule
Model &Compression Method &Compression Setting &LoRA initialization&ARC-C $\uparrow$&MathQA $\uparrow$\\\cmidrule{1-6}
\multirow{8}{*}{LLaMA3-8B} &\multirow{2}{*}{Full-precision} &\multirow{2}{*}{-} &w/o fine-tuning &50.42 &40.10 \\\cmidrule{4-6}
& & &Standard &56.39 &53.56 \\\cmidrule{2-6}
&\multirow{5}{*}{GPTQ} &\multirow{5}{*}{W4} &w/o fine-tuning &45.90 &34.07 \\\cmidrule{4-6}
& & &QLoRA &54.09 &51.42 \\\cmidrule{4-6}
& & &LoftQ &54.52 &53.96 \\\cmidrule{4-6}
& & &EoRA (Ours) &\textbf{55.46} &\textbf{56.04} \\\midrule
\bottomrule
\end{tabular}
}
\end{table}

%% file: tables/appendix_sparse_table.tex
\begin{table}[!htp]\centering
\caption{Perplexity and Commonsense/Math reasoning results of LLaMA2/3 pruned by SparseGPT to 2:4 sparsity, with low-rank compensation of rank 128.}\label{tab:appendix_sparsegpt_r128}
\resizebox{1\columnwidth}{!}{
\begin{tabular}{cccccccc}\toprule
Model &Sparsity &Compensation Method &Wikitext2 $\downarrow$ &ARC-C $\uparrow$&MathQA $\uparrow$&GSM8K $\uparrow$ \\\cmidrule{1-7}
\multirow{9}{*}{LLaMA3-8B} &- &- &6.13 &50.42 &40.10 &36.23 \\\cmidrule{2-7}
&\multirow{7}{*}{2:4} &- &12.32 &30.11 &26.43 &2.12 \\\cmidrule{3-7}
& &ZeroQuant-V2 &11.31 &31.99 &26.49 &2.96 \\\cmidrule{3-7}
& &Act-S &11.32 &31.74 &26.73 &3.26 \\\cmidrule{3-7}
& &ApiQ &\underline{11.08} &\underline{34.21} &\underline{28.77} &\textbf{14.55} \\\cmidrule{3-7}
& &EoRA (Ours) &\textbf{11.07} &\textbf{34.64} &\textbf{29.91} &\underline{13.95} \\\cmidrule{1-7}
\multirow{9}{*}{LLaMA2-7B} &- &- &5.47 &39.84 &27.67 &14.85 \\\cmidrule{2-7}
&\multirow{7}{*}{2:4} &- &8.77 &30.11 &24.65 &1.66 \\\cmidrule{3-7}
& &ZeroQuant-V2 &8.15 &30.54 &24.89 &1.97 \\\cmidrule{3-7}
& &Act-S &8.22 &30.20 &25.09 &2.73 \\\cmidrule{3-7}
& &ApiQ &\underline{8.03} &\underline{32.67} &\textbf{26.36} &\textbf{7.58} \\\cmidrule{3-7}
& &EoRA (Ours) &\textbf{7.97} &\textbf{32.67} &\underline{25.59} &\underline{6.22} \\\cmidrule{1-7}
\multirow{9}{*}{LLaMA2-13B} &- &- &4.88 &45.56 &29.91 &21.37 \\\cmidrule{2-7}
&\multirow{7}{*}{2:4} &- &7.10 &34.30 &25.92 &2.65 \\\cmidrule{3-7}
& &ZeroQuant-V2 &6.82 &33.61 &25.12 &3.56 \\\cmidrule{3-7}
& &Act-S &6.92 &34.12 &25.69 &4.09 \\\cmidrule{3-7}
& &ApiQ &\underline{6.80} &\underline{36.68} &\underline{27.16} &\textbf{12.13} \\\cmidrule{3-7}
& &EoRA (Ours) &\textbf{6.75} &\textbf{37.54} &\textbf{27.53} &\underline{10.91} \\\midrule
\bottomrule
\end{tabular}
}
\end{table}

%% file: tables/appendix_quant_table.tex
\begin{table}[!htp]\centering
\caption{Perplexity and Commonsense/Math reasoning results of LLaMA2/3 quantized by GPTQ with different bit-width, with low-rank compensation of rank 128.}\label{tab:appendix_gptq_r128}
\resizebox{1\columnwidth}{!}{
\begin{tabular}{cccccccc}\toprule
Model &W bits &Compensation Method &Wikitext2 $\downarrow$ &ARC-C $\uparrow$&MathQA $\uparrow$&GSM8K $\uparrow$ \\\cmidrule{1-7}
\multirow{16}{*}{LLaMA3-8B} &- &- &6.13 &50.42 &40.10 &36.23 \\\cmidrule{2-7}
&\multirow{7}{*}{W4} &- &7.00 &45.90 &34.07 &27.74 \\\cmidrule{3-7}
& &ZeroQuant-V2 &\underline{6.80} &45.24 &\underline{36.51} &\textbf{31.23} \\\cmidrule{3-7}
& &Act-S &6.82 &\textbf{47.86} &35.84 &29.34 \\\cmidrule{3-7}
& &ApiQ &6.87 &46.58 &36.18 &30.09 \\\cmidrule{3-7}
& &EoRA (Ours) &\textbf{6.80} &\underline{47.44} &\textbf{37.21} &\underline{30.70} \\\cmidrule{2-7}
&\multirow{7}{*}{W3} &- &15.64 &20.90 &22.37 &0.45 \\\cmidrule{3-7}
& &ZeroQuant-V2 &10.24 &30.02 &26.43 &3.79 \\\cmidrule{3-7}
& &Act-S &\underline{10.19} &\underline{31.28} &25.42 &4.09 \\\cmidrule{3-7}
& &ApiQ &10.41 &30.46 &\underline{26.86} &\underline{10.79} \\\cmidrule{3-7}
& &EoRA (Ours) &\textbf{10.06} &\textbf{31.74} &\textbf{29.11} &\textbf{11.90} \\\cmidrule{1-7}
\multirow{16}{*}{LLaMA2-7B} &- &- &5.47 &39.84 &27.67 &14.85 \\\cmidrule{2-7}
&\multirow{7}{*}{W4} &- &5.75 &38.13 &26.73 &9.93 \\\cmidrule{3-7}
& &ZeroQuant-V2 &\underline{5.68} &37.62 &27.06 &10.15 \\\cmidrule{3-7}
& &Act-S &\underline{5.68} &\textbf{39.84} &\textbf{27.50} &9.86 \\\cmidrule{3-7}
& &ApiQ &\underline{5.68} &\underline{39.59} &27.00 &\underline{11.22} \\\cmidrule{3-7}
& &EoRA (Ours) &\textbf{5.68} &38.05 &\underline{27.13} &\textbf{11.45} \\\cmidrule{2-7}
&\multirow{7}{*}{W3} &- &7.76 &31.65 &23.50 &0.38 \\\cmidrule{3-7}
& &ZeroQuant-V2 &\underline{6.84} &\underline{34.47} &23.90 &2.04 \\\cmidrule{3-7}
& &Act-S &6.86 &32.67 &25.02 &2.57 \\\cmidrule{3-7}
& &ApiQ &6.86 &33.70 &\textbf{26.06} &\underline{7.13} \\\cmidrule{3-7}
& &EoRA (Ours) &\textbf{6.84} &\textbf{35.83} &\underline{25.79} &\textbf{7.50} \\\cmidrule{1-7}
\multirow{16}{*}{LLaMA2-13B} &- &- &4.88 &45.56 &29.91 &21.37 \\\cmidrule{2-7}
&\multirow{7}{*}{W4} &- &5.06 &44.28 &29.10 &21.00 \\\cmidrule{3-7}
& &ZeroQuant-V2 &\underline{5.03} &\underline{44.19} &28.97 &19.48 \\\cmidrule{3-7}
& &Act-S &5.04 &43.60 &\underline{29.48} &18.49 \\\cmidrule{3-7}
& &ApiQ &5.04 &42.83 &\textbf{29.64} &\underline{21.45} \\\cmidrule{3-7}
& &EoRA (Ours) &\textbf{5.03} &\textbf{44.53} &28.90 &\textbf{22.36} \\\cmidrule{2-7}
&\multirow{7}{*}{W3} &- &5.99 &37.28 &26.26 &4.62 \\\cmidrule{3-7}
& &ZeroQuant-V2 &\underline{5.76} &37.54 &26.83 &9.93 \\\cmidrule{3-7}
& &Act-S &5.81 &38.90 &26.26 &9.17 \\\cmidrule{3-7}
& &ApiQ &5.81 &\textbf{39.67} &\textbf{27.47} &\underline{14.32} \\\cmidrule{3-7}
& &EoRA (Ours) &\textbf{5.75} &\underline{39.50} &\underline{27.20} &\textbf{15.08} \\\midrule
\bottomrule
\end{tabular}
}
\end{table}

%% file: tables/appendix_sparse_and_quant_table.tex
\begin{table}[!htp]\centering
\caption{Perplexity and Commonsense/Math reasoning results of LLaMA2/3 models pruned to 2:4 using SparseGPT and quantized to 4-bit with GPTQ, with compensation rank set to 128.}\label{tab:appendix_sparse_and_quant}
\resizebox{1\columnwidth}{!}{
\begin{tabular}{ccccccccc}\toprule
Model &Sparsity &W bits &Compensation Method &Wikitext2 $\downarrow$ &ARC-C $\uparrow$&MathQA $\uparrow$&GSM8K $\uparrow$ \\\cmidrule{1-8}
\multirow{9}{*}{LLaMA3-8B} &- &- &- &6.13 &50.42 &40.10 &36.23 \\\cmidrule{2-8}
&\multirow{7}{*}{2:4} &\multirow{7}{*}{W4} &- &86.15 &18.34 &19.89 &0.00 \\\cmidrule{4-8}
& & &ZeroQuant-V2 &12.84 &29.35 &26.86 &1.59 \\\cmidrule{4-8}
& & &Act-S &12.99 &27.90 &25.59 &1.90 \\\cmidrule{4-8}
& & &ApiQ &\underline{12.77} &\underline{30.71} &\underline{28.74} &\textbf{11.06} \\\cmidrule{4-8}
& & &EoRA (Ours) &\textbf{12.60} &\textbf{31.22} &\textbf{29.58} &\underline{10.16} \\\cmidrule{1-8}
\multirow{9}{*}{LLaMA2-7B} &- &- &- &5.47 &39.84 &27.67 &14.85 \\\cmidrule{2-8}
&\multirow{7}{*}{2:4} &\multirow{7}{*}{W4} &- &9.37 &29.43 &23.88 &0.99 \\\cmidrule{4-8}
& & &ZeroQuant-V2 &8.42 &29.94 &\underline{24.42} &1.67 \\\cmidrule{4-8}
& & &Act-S &\underline{8.24} &28.92 &24.05 &1.97 \\\cmidrule{4-8}
& & &ApiQ &\textbf{8.03} &\underline{30.63} &24.12 &\textbf{7.05} \\\cmidrule{4-8}
& & &EoRA (Ours) &\underline{8.24} &\textbf{31.14} &\textbf{25.39} &\underline{4.93} \\\cmidrule{1-8}
\multirow{9}{*}{LLaMA2-13B} &- &- &- &4.88 &45.56 &29.91 &21.37 \\\cmidrule{2-8}
&\multirow{7}{*}{2:4} &\multirow{7}{*}{W4} &- &7.27 &33.10 &24.75 &2.20 \\\cmidrule{4-8}
& & &ZeroQuant-V2 &6.98 &33.27 &25.29 &2.65 \\\cmidrule{4-8}
& & &Act-S &6.92 &34.64 &26.09 &2.81 \\\cmidrule{4-8}
& & &ApiQ &\textbf{6.80} &\textbf{36.17} &\underline{26.96} &\textbf{12.59} \\\cmidrule{4-8}
& & &EoRA (Ours) &\underline{6.89} &\underline{35.06} &\textbf{27.06} &\underline{9.86} \\\midrule
\bottomrule
\end{tabular}
}
\end{table}

%% file: tables/different_compression.tex
\begin{table}[!htp]\centering
\caption{Comparison between  compensation methods of rank set to 128 on compensating LLaMA3-8B models pruned to 2:4 sparsity with Wanda on Perplexity and Commonsense/Math reasoning tasks.}\label{tab:dif_compression}
\resizebox{1\columnwidth}{!}{
\begin{tabular}{cccccccc}\toprule
Compression Method &Compression Setting &Compensation Method &Wikitext2 $\downarrow$ &ARC-C $\uparrow$&MathQA $\uparrow$&GSM8K $\uparrow$ \\\cmidrule{1-7}
Full-precision &- &- &6.13 &50.42 &40.10 &36.23 \\\cmidrule{1-7}
\multirow{7}{*}{Wanda} &\multirow{7}{*}{2:4} &- &21.42 &27.04 &25.09 &0.76 \\\cmidrule{3-7}
& &ZeroQuant-V2 &17.16 &30.46 &26.16 &1.28 \\\cmidrule{3-7}
& &Act-S &17.37 &29.77 &26.73 &1.51 \\\cmidrule{3-7}
& &ApiQ &\underline{14.30} &\underline{31.91} &\underline{29.61} &\textbf{12.81} \\\cmidrule{3-7}
& &EoRA (Ours) &\textbf{14.04} &\textbf{34.81} &\textbf{30.05} &\underline{11.52} \\\midrule
\bottomrule
\end{tabular}
}
\end{table}

%% file: tables/appendix_calib.tex
\begin{table}[!htp]\centering
\caption{Ablation studies of calibrating EoRA with different calibration sizes on compensating compressed LLaMA3-8B.}\label{tab:appendix_calib}
\resizebox{0.7\columnwidth}{!}{
\begin{tabular}{ccccc}\toprule
Model &Quantization Format &\#Calib. &Calib. Time (mins) &MathQA $\uparrow$ \\\cmidrule{1-5}
\multirow{17}{*}{LLaMA3-8B} &FP16 &- &- & 40.10 \\\cmidrule{2-5}
&\multirow{8}{*}{W4} &16 &6.40 &36.62 \\\cmidrule{3-5}
& &32 &7.04 &36.93 \\\cmidrule{3-5}
& &64 &8.03 &37.21 \\\cmidrule{3-5}
& &128 &10.40 &37.46 \\\cmidrule{3-5}
& &256 &14.43 &37.60 \\\cmidrule{3-5}
& &512 &21.11 &37.30 \\\cmidrule{2-5}
&\multirow{8}{*}{W3} &16 &6.33 &26.33 \\\cmidrule{3-5}
& &32 &7.20 &27.57 \\\cmidrule{3-5}
& &64 &8.16 &29.11 \\\cmidrule{3-5}
& &128 &11.33 &30.34 \\\cmidrule{3-5}
& &256 &14.17 &30.21 \\\cmidrule{3-5}
& &512 &20.89 &30.40 \\\midrule
\bottomrule
\end{tabular}
}
\end{table}

%% file: tables/speedup.tex
\begin{table}[!htp]\centering
\caption{Comparison of the average per-token latency (batch size 1) for 128-token generation on LLaMA3-70B between full-precision and GPTQ + EoRA with and without our custom EoRA kernel.}\label{tab:speedup}
\resizebox{0.6\columnwidth}{!}{%
\begin{tabular}{lcccc}\toprule
Format &EoRA Rank &EoRA Kernel &Latency $\downarrow$ &Speedup $\uparrow$ \\\cmidrule{1-5}
FP-16 &- &- &60ms &1x \\\cmidrule{1-5}
\multirow{9}{*}{3-bit} &- &- &35ms &1.7x \\\cmidrule{2-5}
&64 & {\color{Red}No} & {\color{Red}52ms} & {\color{Red}1.2x} \\\cmidrule{2-5}
&64 &{\color{ForestGreen}Yes} & {\color{ForestGreen}44ms} & {\color{ForestGreen} 1.4x}
\\\cmidrule{2-5}
&128 & {\color{Red}No} & {\color{Red}54ms} & {\color{Red}1.1x} \\\cmidrule{2-5}
&128 &{\color{ForestGreen}Yes} & {\color{ForestGreen}43ms} & {\color{ForestGreen} 1.4x} \\\cmidrule{2-5}
&256 & {\color{Red}No} & {\color{Red}58ms} & {\color{Red}1x} \\\cmidrule{2-5}
&256 &{\color{ForestGreen}Yes} & {\color{ForestGreen}48ms} & {\color{ForestGreen} 1.3x} \\\cmidrule{1-5}
\multirow{9}{*}{4-bit} &- &- &38ms &1.6x \\\cmidrule{2-5}
&64 & {\color{Red}No} & {\color{Red}60ms} & {\color{Red}1x} \\\cmidrule{2-5}
&64 &{\color{ForestGreen}Yes} & {\color{ForestGreen}49ms} & {\color{ForestGreen} 1.2x}
\\\cmidrule{2-5}
&128 & {\color{Red}No} & {\color{Red}61ms} & {\color{Red}1x} \\\cmidrule{2-5}
&128 & {\color{ForestGreen}Yes} & {\color{ForestGreen}51ms} & {\color{ForestGreen}1.2x} \\\cmidrule{2-5}
&256 & {\color{Red}No} & {\color{Red}63ms} & {\color{Red}1x} \\\cmidrule{2-5}
&256 &{\color{ForestGreen}Yes} & {\color{ForestGreen}53ms} & {\color{ForestGreen} 1.1x} \\\cmidrule{1-5}
\bottomrule
\end{tabular}
}
\end{table}

%% file: tables/appendix_different_rank.tex
\begin{table}[!htp]\centering
\caption{Results of \methodabbr~of different rank on compensating LLaMA2/3 models pruned to 2:4 sparsity by SparseGPT on Commonsense and Math reasoning tasks.}\label{tab:appendix_diff_rank}
\resizebox{0.58\columnwidth}{!}{%
\begin{tabular}{cccccccc}\toprule
Model &Sparsity &r &Compensation Method &ARC-C $\uparrow$&MathQA $\uparrow$&GSM8K $\uparrow$ \\\cmidrule{1-7}
\multirow{26}{*}{LLaMA3-8B} &- &- &- &50.42 &40.10 &36.23 \\\cmidrule{2-7}
&\multirow{24}{*}{2:4} &- &- &30.11 &26.43 &2.12 \\\cmidrule{3-7}
& &\multirow{5}{*}{64} &ZeroQuant-V2 &30.97 &26.39 &2.27 \\\cmidrule{4-7}
& & &Act-S &30.46 &26.67 &3.34 \\\cmidrule{4-7}
& & &ApiQ &\underline{33.10} &\underline{27.87} &\textbf{11.52} \\\cmidrule{4-7}
& & &EoRA (Ours) &\textbf{33.10} &\textbf{28.57} &\underline{10.77} \\\cmidrule{3-7}
& &\multirow{5}{*}{128} &ZeroQuant-V2 &31.99 &26.49 &2.96 \\\cmidrule{4-7}
& & &Act-S &31.74 &26.73 &3.26 \\\cmidrule{4-7}
& & &ApiQ &\underline{34.21} &\underline{28.77} &\textbf{14.55} \\\cmidrule{4-7}
& & &EoRA (Ours) &\textbf{34.64} &\textbf{29.91} &\underline{13.95} \\\cmidrule{3-7}
& &\multirow{5}{*}{256} &ZeroQuant-V2 &34.55 &28.74 &4.09 \\\cmidrule{4-7}
& & &Act-S &32.76 &27.94 &5.16 \\\cmidrule{4-7}
& & &ApiQ &\underline{35.41} &\underline{30.45} &\underline{15.85} \\\cmidrule{4-7}
& & &EoRA (Ours) &\textbf{37.96} &\textbf{31.59} &\textbf{17.06} \\\cmidrule{3-7}
& &\multirow{5}{*}{512} &ZeroQuant-V2 &\underline{38.73} &30.38 &6.75 \\\cmidrule{4-7}
& & &Act-S &36.18 &29.65 &8.64 \\\cmidrule{4-7}
& & &ApiQ &36.69 &\underline{32.63} &\underline{20.77} \\\cmidrule{4-7}
& & &EoRA (Ours) &\textbf{41.89} &\textbf{34.17} &\textbf{23.28} \\\cmidrule{1-7}
\multirow{26}{*}{LLaMA2-7B} &- &- &- &39.84 &27.67 &14.85 \\\cmidrule{2-7}
&\multirow{24}{*}{2:4} &- &- &30.11 &24.65 &1.66 \\\cmidrule{3-7}
& &\multirow{5}{*}{64} &ZeroQuant-V2 &30.20 &24.48 &1.97 \\\cmidrule{4-7}
& & &Act-S &30.12 &25.03 &1.74 \\\cmidrule{4-7}
& & &ApiQ &\underline{31.83} &\underline{25.62} &\textbf{5.91} \\\cmidrule{4-7}
& & &EoRA (Ours) &\textbf{32.16} &\textbf{25.62} &\underline{5.08} \\\cmidrule{3-7}
& &\multirow{5}{*}{128} &ZeroQuant-V2 &30.54 &24.89 &1.97 \\\cmidrule{4-7}
& & &Act-S &30.20 &25.09 &2.73 \\\cmidrule{4-7}
& & &ApiQ &\underline{32.67} &\textbf{26.36} &\textbf{7.58} \\\cmidrule{4-7}
& & &EoRA (Ours) &\textbf{32.67} &\underline{25.59} &\underline{6.22} \\\cmidrule{3-7}
& &\multirow{5}{*}{256} &ZeroQuant-V2 &31.99 &25.19 &2.88 \\\cmidrule{4-7}
& & &Act-S &32.59 &25.39 &3.26 \\\cmidrule{4-7}
& & &ApiQ &\underline{34.30} &\underline{25.99} &\textbf{8.79} \\\cmidrule{4-7}
& & &EoRA (Ours) &\textbf{34.47} &\textbf{26.06} &\underline{7.88} \\\cmidrule{3-7}
& &\multirow{5}{*}{512} &ZeroQuant-V2 &34.72 &24.38 &3.34 \\\cmidrule{4-7}
& & &Act-S &34.73 &25.76 &3.56 \\\cmidrule{4-7}
& & &ApiQ &\underline{34.98} &\textbf{26.16} &\textbf{9.70} \\\cmidrule{4-7}
& & &EoRA (Ours) &\textbf{36.77} &\underline{25.96} &\underline{8.79} \\\cmidrule{1-7}
\multirow{26}{*}{LLaMA2-13B} &- &- &- &45.56 &29.91 &21.37 \\\cmidrule{2-7}
&\multirow{24}{*}{2:4} &- &- &34.30 &25.92 &2.65 \\\cmidrule{3-7}
& &\multirow{5}{*}{64} &ZeroQuant-V2 &33.95 &25.56 &2.81 \\\cmidrule{4-7}
& & &Act-S &32.76 &25.93 &2.96 \\\cmidrule{4-7}
& & &ApiQ &\underline{35.84} &\textbf{27.17} &\textbf{8.64} \\\cmidrule{4-7}
& & &EoRA (Ours) &\textbf{36.00} &\underline{26.80} &\underline{8.19} \\\cmidrule{3-7}
& &\multirow{5}{*}{128} &ZeroQuant-V2 &33.61 &25.12 &3.56 \\\cmidrule{4-7}
& & &Act-S &34.12 &25.69 &4.09 \\\cmidrule{4-7}
& & &ApiQ &\underline{36.68} &\underline{27.16} &\textbf{12.13} \\\cmidrule{4-7}
& & &EoRA (Ours) &\textbf{37.54} &\textbf{27.53} &\underline{10.91} \\\cmidrule{3-7}
& &\multirow{5}{*}{256} &ZeroQuant-V2 &35.06 &26.06 &4.93 \\\cmidrule{4-7}
& & &Act-S &34.56 &26.23 &4.62 \\\cmidrule{4-7}
& & &ApiQ &\underline{36.69} &\underline{27.40} &\textbf{14.56} \\\cmidrule{4-7}
& & &EoRA (Ours) &\textbf{38.73} &\textbf{27.77} &\underline{13.04} \\\cmidrule{3-7}
& &\multirow{5}{*}{512} &ZeroQuant-V2 &36.51 &26.39 &7.28 \\\cmidrule{4-7}
& & &Act-S &36.86 &26.77 &6.14 \\\cmidrule{4-7}
& & &ApiQ &\underline{38.57} &\underline{27.71} &\underline{17.21} \\\cmidrule{4-7}
& & &EoRA (Ours) &\textbf{40.61} &\textbf{29.17} &\textbf{17.51} \\\midrule
\bottomrule
\end{tabular}
}
\end{table}

%% file: tables/appendix_finetuning.tex
\begin{table}[!htp]\centering
\caption{Fine-tune the compressed LLaMA3-8B models with various compression settings and different initialization of the low-rank matrices for Commonsense/Math reasoning tasks.}\label{tab:appendix_finetuning}
\resizebox{1\columnwidth}{!}{
\begin{tabular}{ccccccc}\toprule
Model &Compression Method &Compression Setting &LoRA initialization &ARC-C $\uparrow$&MathQA $\uparrow$ \\\cmidrule{1-6}
\multirow{20}{*}{LLaMA3-8B} &\multirow{2}{*}{Full-precision} &\multirow{2}{*}{-} &w/o fine-tuning &50.42 &40.10 \\\cmidrule{4-6}
& & &Standard &56.39 &53.56 \\\cmidrule{2-6}
&\multirow{6}{*}{SparseGPT} &\multirow{6}{*}{2:4} &w/o fine-tuning &30.11 &26.43 \\\cmidrule{4-6}
& & &QLoRA &41.30 &45.42 \\\cmidrule{4-6}
& & &LoftQ &43.68 &48.77 \\\cmidrule{4-6}
& & &EoRA (Ours) &\textbf{48.54} &\textbf{54.67} \\\cmidrule{2-6}
&\multirow{6}{*}{GPTQ} &\multirow{6}{*}{W4} &w/o fine-tuning &45.90 &34.07 \\\cmidrule{4-6}
& & &QLoRA &54.09 &51.42 \\\cmidrule{4-6}
& & &LoftQ &54.52 &53.96 \\\cmidrule{4-6}
& & &EoRA (Ours) &\textbf{55.46} &\textbf{56.04} \\\cmidrule{2-6}
&\multirow{6}{*}{GPTQ} &\multirow{6}{*}{W3} &w/o fine-tuning &20.90 &22.37 \\\cmidrule{4-6}
& & &QLoRA &30.29 &34.10 \\\cmidrule{4-6}
& & &LoftQ &44.70 &48.17 \\\cmidrule{4-6}
& & &EoRA (Ours) &\textbf{47.44} &\textbf{53.90} \\\midrule
\bottomrule
\end{tabular}
}
\end{table}

%% file: tables/finetuning_diff_dataset_ratio_table.tex
\begin{table}[h]\centering
\caption{Ablation study on the effect of using different proportions of the dataset for fine-tuning 2:4 pruned LLaMA3-8B models with varying low-rank matrix initializations on Commonsense/Math reasoning tasks.}\label{tab:finetuning_diff_dataset_ratio}
\resizebox{0.7\columnwidth}{!}{
\begin{tabular}{lcccc}\toprule
Model &Dataset Ratio &LoRA initialization &ARC-C $\uparrow$ &MathQA $\uparrow$ \\\cmidrule{1-5}
\multirow{14}{*}{LLaMA3-8B} &- &- &50.42 &40.10 \\\cmidrule{2-5}
&\multirow{3}{*}{100\%} &QLoRA &41.30 &45.42 \\\cmidrule{3-5}
& &LoftQ &43.68 &48.77 \\\cmidrule{3-5}
& &EoRA (Ours) &\textbf{48.54} &\textbf{54.67} \\\cmidrule{2-5}
&\multirow{3}{*}{50\%} &QLoRA &38.56 &40.23 \\\cmidrule{3-5}
& &LoftQ &41.46 &42.51 \\\cmidrule{3-5}
& &EoRA (Ours) &\textbf{46.41} &\textbf{48.91} \\\cmidrule{2-5}
&\multirow{3}{*}{30\%} &QLoRA &36.77 &36.71 \\\cmidrule{3-5}
& &LoftQ &39.76 &40.60 \\\cmidrule{3-5}
& &EoRA (Ours) &\textbf{43.85} &\textbf{44.79} \\\midrule
\bottomrule
\end{tabular}
}
\end{table}

%% file: tables/appendix_lowrank_quantization.tex
\begin{table}[ht]\centering
\caption{Accuracy and the Model Size of quantizing \methodabbr~of rank \{128,512\} to 4/3-bit on compensating LLaMA3-8B of \{2:4 sparisity, 4/3-bit\}.}
\label{tab:appendix_quant_eora}
\resizebox{1\columnwidth}{!}{
\begin{tabular}{cccccccc}\toprule
Compression method & Config &r &W-bit of EoRA &Model Size (GB) &ARC-C $\uparrow$&MathQA $\uparrow$ \\\cmidrule{1-7}
- &- &- &- &15.08 &50.42 &40.10 \\\cmidrule{1-7}
\multirow{10}{*}{SparseGPT} &\multirow{10}{*}{2:4} &- &- &9.12 &30.11 &26.43 \\\cmidrule{3-7}
& &\multirow{4}{*}{128} &16 &9.77 &34.64 &29.91 \\\cmidrule{4-7}
& & &4 &9.28 &34.47 &29.91 \\\cmidrule{4-7}
& & &3 &9.24 &34.72 &29.71 \\\cmidrule{3-7}
& &\multirow{4}{*}{512} &16 &11.70 &41.89 &34.17 \\\cmidrule{4-7}
& & &4 &9.77 &41.46 &33.63 \\\cmidrule{4-7}
& & &3 &9.64 &40.35 &32.66 \\\cmidrule{1-7}
\multirow{22}{*}{GPTQ} &\multirow{10}{*}{W4} &- &- &5.35 &45.90 &34.07 \\\cmidrule{3-7}
& &\multirow{4}{*}{128} &16 &6.01 &47.44 &37.21 \\\cmidrule{4-7}
& & &4 &5.50 &47.35 &36.78 \\\cmidrule{4-7}
& & &3 &5.46 &47.18 &36.52 \\\cmidrule{3-7}
& &\multirow{4}{*}{512} &16 &7.85 &48.29 &38.72 \\\cmidrule{4-7}
& & &4 &6.01 &48.80 &38.92 \\\cmidrule{4-7}
& & &3 &5.90 &46.92 &36.88 \\\cmidrule{2-7}
&\multirow{10}{*}{W3} &- &- &4.63 &20.90 &22.37 \\\cmidrule{3-7}
& &\multirow{4}{*}{128} &16 &5.28 &31.74 &29.11 \\\cmidrule{4-7}
& & &4 &4.78 &31.48 &28.64 \\\cmidrule{4-7}
& & &3 &4.74 &29.18 &26.7 \\\cmidrule{3-7}
& &\multirow{4}{*}{512} &16 &7.16 &38.82 &31.89 \\\cmidrule{4-7}
& & &4 &5.28 &40.01 &31.69 \\\cmidrule{4-7}
& & &3 &5.18 &35.4 &30.45 \\\midrule
\bottomrule
\end{tabular}
}
\end{table}

%% file: tables/llm_summarization.tex
\begin{table}[!htp]\centering
\caption{Comparison of 4-bit and 3-bit quantized LLaMA3-8B on the LLM summarization task.}\label{tab:llm_summarization}
\resizebox{0.9\columnwidth}{!}{
\begin{tabular}{ccccc}\toprule
Model &Quantization Format &Compensation Method &CNN/DailyMail (ROUGE-Lsum) $\uparrow$ \\\cmidrule{1-4}
\multirow{13}{*}{LLaMA3-8B} &\multirow{7}{*}{4-bit} &- &0.1672 \\\cmidrule{3-4}
& &ZeroQuant-V2 &0.1798 \\\cmidrule{3-4}
& &Act-S &0.1786 \\\cmidrule{3-4}
& &ApiQ &0.1804 \\\cmidrule{3-4}
& &EoRA (Ours) &\textbf{0.1812} \\\cmidrule{2-4}
&\multirow{7}{*}{3-bit} &- &0.0650 \\\cmidrule{3-4}
& &ZeroQuant-V2 &0.0970 \\\cmidrule{3-4}
& &Act-S &0.1286 \\\cmidrule{3-4}
& &ApiQ &0.1357 \\\cmidrule{3-4}
& &EoRA (Ours) &\textbf{0.1463}  \\\midrule
\bottomrule
\end{tabular}
}
\end{table}

%% file: tables/mmlu.tex
\begin{table}[!htp]\centering
\caption{Comparison of 4-bit and 3-bit quantized LLaMA3.2-3B on multi-task language understanding.}\label{tab:mmlu}
\resizebox{0.7\columnwidth}{!}{
\begin{tabular}{ccccc}\toprule
Model &Quantization Format &EoRA Rank &MMLU $\uparrow$ \\\cmidrule{1-4}
\multirow{13}{*}{LLaMA3.2-3B} &FP16 &- &54.19 \\\cmidrule{2-4}
&\multirow{5}{*}{4-bit} &- &24.16 \\\cmidrule{3-4}
& &32 &52.53 \\\cmidrule{3-4}
& &64 &52.49 \\\cmidrule{3-4}
& &128 &52.93 \\\cmidrule{2-4}
&\multirow{5}{*}{3-bit} &- &22.89 \\\cmidrule{3-4}
& &32 &39.08 \\\cmidrule{3-4}
& &64 &38.83 \\\cmidrule{3-4}
& &128 &39.68  \\\midrule
\bottomrule
\end{tabular}
}
\end{table}

%% file: tables/groupwise_eora.tex
\begin{table}[!htp]\centering
\caption{Comparison of 4-bit and 3-bit group-wise quantized LLaMA3-8B. The group size is set as 128. EoRA rank is set as 128.}\label{tab:groupwise}
\resizebox{0.7\columnwidth}{!}{
\begin{tabular}{ccccc}\toprule
Model &Quantization Format &Compensation Method &MathQA $\uparrow$ \\\cmidrule{1-4}
\multirow{15}{*}{LLaMA3-8B} &FP16 &- &40.10 \\\cmidrule{2-4}
&\multirow{7}{*}{W4-groupsize 128} &- & 38.34 \\\cmidrule{3-4}
& &ZeroQuant-V2 &38.92 \\\cmidrule{3-4}
& &Act-S &38.49 \\\cmidrule{3-4}
& &ApiQ &38.90 \\\cmidrule{3-4}
& &EoRA (Ours) &\textbf{39.16} \\\cmidrule{2-4}
&\multirow{7}{*}{W3-groupsize 128} &- &32.52 \\\cmidrule{3-4}
& &ZeroQuant-V2 &32.39 \\\cmidrule{3-4}
& &Act-S &33.33 \\\cmidrule{3-4}
& &ApiQ &34.80 \\\cmidrule{3-4}
& &EoRA (Ours) &\textbf{35.10}  \\\midrule
\bottomrule
\end{tabular}
}
\end{table}

%% file: tables/layerwise_discrepancy.tex
\begin{table}[!htp]\centering
\caption{Comparison of layer-wise discrepancy (the less the better) on $q$ projector of LLaMA2-7B.}
\label{tab:layerwise_discrepancy}
\resizebox{0.8\columnwidth}{!}{
\begin{tabular}{cccccccc}\toprule
Method & Layer 0 & Layer 5 & Layer 10 & Layer 15 & Layer 20 & Layer 25 & Layer 30 \\\midrule
GPTQ (4-bit) & 2.3 & 11.3 & 13.4 & 12.6 & 10.7 & 10.9 & 11.5 \\\midrule
ZeroQuant-V2 & 1.7 & 10.2 & 10.8 & 11.3 & 9.7 & 9.2 & 9.9 \\\midrule
Act-S & 2.0 & 9.7 & 9.4 & 10.3 & 8.4 & 7.7 & 8.9 \\\midrule
ApiQ & 1.5 & 7.8 & 8.2 & 7.4 & 6.2 & 5.9 & 5.1 \\\midrule
EoRA (Ours) & 1.4 & 8.5 & 8.1 & 7.2 & 5.8 & 6.0 & 5.4 \\\midrule
\bottomrule
\end{tabular}
}
\end{table}

%% file: tables/more_comparison.tex
\begin{table}[!htp]\centering
\caption{Comparison of 4-bit and 3-bit quantized LLaMA3-8B for more recent low-rank approaches.}\label{tab:more_comparison}
\resizebox{0.7\columnwidth}{!}{
\begin{tabular}{ccccc}\toprule
Model &Quantization Format &Compensation Method &MathQA $\uparrow$ \\\cmidrule{1-4}
\multirow{36}{*}{LLaMA3-8B} &- &- &40.10 \\\cmidrule{2-4}
&\multirow{16}{*}{4-bit} &- &34.07 \\\cmidrule{3-4}
& &FWSVD &35.64 \\\cmidrule{3-4}
& &ZeroQuant-V2 &36.51 \\\cmidrule{3-4}
& &Act-S &35.84 \\\cmidrule{3-4}
& &ApiQ &36.18 \\\cmidrule{3-4}
& &LQER &35.46 \\\cmidrule{3-4}
& &LRC &36.40 \\\cmidrule{3-4}
& &CALDERA &36.70 \\\cmidrule{3-4}
& &QERA &35.90 \\\cmidrule{3-4}
& &SLiM &35.90 \\\cmidrule{3-4}
& &OATS &36.01 \\\cmidrule{3-4}
& &EoRA (Ours) &\textbf{37.21} \\\cmidrule{2-4}
&\multirow{16}{*}{3-bit} &- &22.37 \\\cmidrule{3-4}
& &FWSVD &26.30 \\\cmidrule{3-4}
& &ZeroQuant-V2 &26.43 \\\cmidrule{3-4}
& &Act-S &25.42 \\\cmidrule{3-4}
& &ApiQ &26.86 \\\cmidrule{3-4}
& &LQER &25.60 \\\cmidrule{3-4}
& &LRC &28.64 \\\cmidrule{3-4}
& &CALDERA &28.10 \\\cmidrule{3-4}
& &QERA &25.32 \\\cmidrule{3-4}
& &SLiM &25.91 \\\cmidrule{3-4}
& &OATS &25.30 \\\cmidrule{3-4}
& &EoRA (Ours) &\textbf{29.11}  \\\midrule
\bottomrule
\end{tabular}
}
\end{table}

%% file: tables/2bit_gptq.tex
\begin{table}[!htp]\centering
\caption{Comparison of 2-bit GPTQ-quantized LLaMA3-8B on MathQA.}\label{tab:2bit_gptq}
\resizebox{0.8\columnwidth}{!}{
\begin{tabular}{ccccc}\toprule
Model &Quantization Format &Fine-tuning Strategy &MathQA $\uparrow$ \\\cmidrule{1-4}
\multirow{7}{*}{LLaMA3-8B} &\multirow{7}{*}{2-bit (GPTQ)} &- &18.22 \\\cmidrule{3-4}
& &LoftQ &35.80 \\\cmidrule{3-4}
& &EoRA &36.89 \\\cmidrule{3-4}
& &RILQ &37.60 \\\cmidrule{3-4}
& &RILQ + EoRA &\textbf{38.90} \\\midrule
\bottomrule
\end{tabular}
}
\end{table}